\documentclass[fleqn,10pt]{article}

\usepackage[utf8]{inputenc}
\usepackage[T1]{fontenc}
\usepackage[a4paper, top=1in, bottom=1in, left=1in, right=1in]{geometry}

\usepackage{tikz}
\usetikzlibrary{shapes.geometric, arrows}
\usepackage{lineno}
\usepackage{tcolorbox}
\usepackage{fontawesome}
\usepackage{graphicx}
\usepackage{adjustbox}
\usepackage{subcaption}
\usepackage{multirow}
\usepackage{url}
\usepackage{amsmath}
\usepackage{array}
\usepackage{xcolor}
\usepackage{booktabs}
\usepackage{authblk}
\usepackage{hyperref}
\newcommand{\add}[1]{#1}
\newcommand{\delete}[1]{}
 \title{
A Global Dataset of Location Data Integrity-Assessed Reforestation Efforts}
\author[1]{Angela John$^{\dag,*}$}
\author[1]{Selvyn Allotey}
\author[1]{Till Koebe}
\author[2]{Alexandra Tyukavina}
\author[1]{Ingmar Weber}

\affil[1]{Saarland Informatics Campus, Department of Computer Science, Saarbrücken, 66123, Germany}
\affil[2]{University of Maryland, Department of Geographical Sciences, Riverdale, 20737, USA}
\affil[]{%
 $^{\dag,*}$ Corresponding author: \texttt{ajohn@cs.uni-saarland.de}
}

\begin{document}

\maketitle
\begin{abstract}

Afforestation and reforestation are popular strategies for mitigating climate change by enhancing carbon sequestration. However, the effectiveness of these efforts is often self-reported by project developers, or certified through processes with limited external validation. This leads to concerns about data reliability and project integrity. In response to increasing scrutiny of voluntary carbon markets, this study presents a dataset on global afforestation and reforestation efforts compiled from primary (meta-)information and augmented with time-series satellite imagery and other secondary data. Our dataset covers 1,289,068 planting sites from 45,628 projects spanning 33 years. Since any remote sensing-based validation effort relies on the integrity of a planting site's geographic boundary, this dataset introduces a standardized assessment of the provided site-level location information, which we summarize in one easy-to-communicate key indicator: LDIS -- the Location Data Integrity Score. We find that approximately 79\% of the georeferenced planting sites monitored fail on at least 1 out of 10 LDIS indicators, while 15\% of the monitored projects lack machine-readable georeferenced data in the first place. In addition to enhancing accountability in the voluntary carbon market, the presented dataset also holds value as training data for e.g. computer vision-related tasks with millions of linked Sentinel-2 and Planetscope satellite images.

\end{abstract}



\section*{Background \& Summary}\label{sec:background}

With forests being one of the major carbon sinks on our planet, there have been numerous efforts to protect, restore and extend forested areas to mitigate climate change \cite{fao2020, rytter2020carbon}.However, while economic rent seeking  through depletion of natural resources such as unsustainable logging, overgrazing, conversion to farmland,establishment of monoculture plantations, and mismanaged firescan limit the effectiveness of these efforts, forests also offer renewable resources (e.g., sustainably managed timber), highlighting the importance of management practices that balance resource use with longterm ecosystem integrity\cite{alados2022two, russo2022understanding}. The rise of the voluntary carbon market (VCM) has provided landowners with an alternative source of income that aligns restoration goals with economic incentives. In contrast to heavily regulated compliance carbon markets such as the European Union Emissions Trading System, VCM prices per ton of CO\textsubscript{2} offset are mainly determined by market forces of supply and demand.
Public controversies that have sparked in recent years around the additionality (see box `Key Terms') of VCM activities, especially of REDD+ projects, have depressed buyer confidence, reducing demand and market volumes in parts of the VCM since 2021. Moreover, recent analyses \cite{west2020overstated, west2023action} demonstrate that many REDD+ projects overstate additionality and fail to deliver the claimed emission reductions, further undermining buyer confidence and contributing to price declines \cite{ecosystemmarketplace2021, ccarbon2023, carbonpulse2023, probst2024systematic}.

\thispagestyle{empty}
\begin{tcolorbox}[colback=blue!5!white, colframe=black!75!black, 
    title=Key Terms, fonttitle=\bfseries, 
    boxrule=0.5mm, rounded corners, width=\textwidth]

    \begin{enumerate}
       
        \item   \textbf{Article 6 of the Paris Agreement:} The Paris Agreement, adopted by the Parties to the United Nations Framework Convention on Climate Change (UNFCCC), aims to limit global temperature rise to  below 2°C above pre-industrial levels, with efforts to keep it to 1.5°C. It emphasizes equity and common but differentiated responsibilities to support sustainable development and poverty eradication. It also promotes the conservation of carbon sinks, such as forests, and encourages actions to reduce emissions from deforestation. Countries must submit their nationally determined contributions (NDCs) every five years, subject to expert review. The Agreement came into force on November 4, 2016, and by April 2021, it had been adopted by 194 Parties.

        \item   \textbf{REDD+:} A UNFCCC-developed voluntary climate change mitigation framework aimed to facilitate developing countries in "Reducing Emissions from Deforestation and forest Degradation”, with "+" emphasizing the role of conservation, sustainable management of forests, and enhancement of forest carbon stocks.  REDD+ can be implemented at multiple levels, from specific project areas (e.g., a forest concession or protected area) to broader subnational or national initiatives.

        \item \textbf{Voluntary carbon markets:} A voluntary carbon market (VCM) is a decentralised platform where private entities can buy and sell carbon credits that represent removals or reductions of greenhouse gases (GHGs) in the atmosphere in avoluntarily beyond regulatory compliance frameworks.
        
        \item   \textbf{Permanence:} 
        In the context of carbon projects, permanence refers to the durability of stored carbon while acknowledging the risk of reversal; it is not a binary attribute but a continuum managed over time through measures such as long-term monitoring, insurance, and other risk mitigation strategies to maintain carbon benefits over extended periods.
        \item   \textbf{Additionality:} The requirement that the emissions reductions or carbon sequestration achieved by a REDD+ or carbon offset project are greater than what would have occurred in the absence of the project.

        \item   \textbf{Verification:} The independent assessment and confirmation of the emissions reductions or carbon sequestration claimed by a carbon offset project, carried out by third-party auditors who evaluate whether the project’s reported outcomes are accurate, credible, and consistent with the standards set by a given carbon market protocol.

       \item \textbf{Deforestation:} Deforestation describes the process of clearing forest from an area and converting the area to non forest(agriculture,bare land,infrastructure etc.).
    
       \item \textbf{Reforestation:} Reforestation describes the process of replanting trees in areas where forests have decreased previously (within the past 50-100 years) due to \delete{depleted} cleared or degraded ecosystems usually through deforestation.
    
       \item \textbf{Afforestation:} The process of planting trees or sowing seeds of trees in areas where previously no trees existed within the past 50-100 years.When applied to degraded or deforested lands, afforestation can provide carbon sequestration and ecosystem benefits; however, planting trees in naturally non‐forested ecosystems (e.g., native grasslands, savannas, peatlands) can cause substantial biodiversity loss and alter ecosystem functions. Thus, site selection and ecosystem context are critical when considering afforestation.
         \item  \textbf{Forest:} According to Food and Agriculture Organization of the United Nations \cite{FAO2000FRA}, "a forest is an area of land of more than 0.05 hectares with a tree canopy cover of at least 10\% with a minimum height between two and five meters. In our work, we use GLAD forest with the forest definition extended to include the trees outside the forest, such as those in agroforestry areas"
    \end{enumerate}
\end{tcolorbox}

A major limitation of current reforestation initiatives is the lack of independent, high-quality spatial data to verify a project's location data integrity. Here, by `integrity' we mean the validity, accuracy and completeness of the spatial information—ensuring that the site’s mapped boundaries are topologically valid (no self‑intersections), free of spurious overlaps or gaps, and closely correspond to the true extent of the on‑the‑ground reforestation efforts. Many datasets rely on self-reported information from project developers or certification bodies, which often lacks external validation. Certification frameworks such as Verra (\url{https://registry.verra.org/}), Gold Standard 
 (\url{https://www.goldstandard.org/}),and the American Carbon Registry (\url{https://wppremiumplugins.com/americancarbonregistry/})
offer structured methodologies but  substantially vary in rigor, leading to inconsistencies in credit quality \cite{offsetguide2020}. These inconsistencies complicate the comparison and aggregation of carbon credits, making it difficult for buyers to assess and ensure the integrity of their offsets. As the implementation of Article 6 of the Paris Agreement (see box `Key Terms') advances, ensuring transparency and standardization in VCM becomes increasingly critical. The major carbon crediting companies have recently responded by joining global compliance efforts such as the United Nations' International Civil Aviation Organization (ICAO) CORSIA\cite{prussi2021corsia} or the Core Carbon Principles (CCPs)\cite{icvcm2024core}, but there is still a long way to go\cite{probst2024systematic}.

To address these challenges, this study introduces a global dataset of quality-assessed, georeferenced reforestation projects. The dataset consolidates information from over 50 sources, covering 1.29 million planting sites from 45,628 projects spanning 33 years. In addition to the core location and project metadata, we augment information about each reforestation site with periodic Sentinel-2 satellite imagery and with secondary datasets capturing infrastructure presence, land-cover transitions, and local climatic conditions. We also use large-language models to extract contextual details directly from project descriptions and accompanying project documentation. Finally, we derive integrity indicators from these variables to flag potential issues with the reported site locations.

The list of websites from which the data are acquired can be found in Table \ref{tab:list_websites} in the Appendix. An overview of the indicators available in the dataset can be found in Table \ref{tab:integrity_indicators} and in Tables \ref{tab:metadata_derived} and \ref{tab:secondary_data} in the Appendix. Figure 1 describes the underlying data aggregation process.\\

\begin{figure}[htbp]
\centering
\includegraphics[width=1\textwidth]{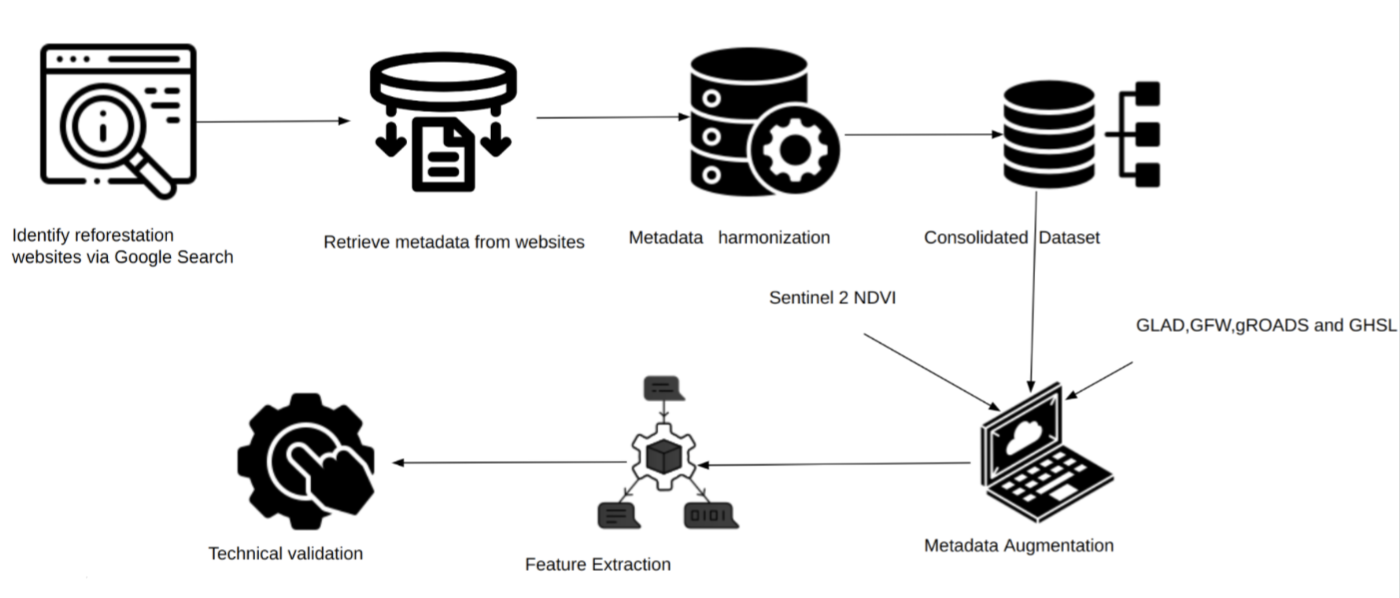}
\caption{The Reforestation Data Generation Workflow}
\label{fig:Flowchart}
\end{figure}

Assessing the success of reforestation efforts is inherently multidimensional, encompassing ecological, social, and economic factors \cite{martin2021people, locatelli2015tropical, fargione2021challenges, kemppinen2020global}. Various indicators have been proposed to evaluate different aspects of success: establishment success (e.g. seedling survival rates, initial tree growth), forest growth success (e.g., stand density, biomass accumulation), environmental success (e.g., biodiversity restoration, soil stabilization, carbon sequestration) and socioeconomic success (e.g., employment generation, community engagement, land tenure security). However, these indicators are rarely standardized across projects and many rely on self-reported data with limited independent verification \cite{le2012more}. Remote sensing offers a scalable approach to monitoring tree cover change \cite{hansen2013high, du2023mapping, potapov2022global, brandt2023wall, brandt2020unexpectedly, tucker2023sub, yao2021tree, velasquez2023implementing}, yet it struggles to capture more nuanced success metrics such as species composition, ecosystem resilience, or socioeconomic benefits \cite{aziz2024remote, dixit2024potential, ecke2022uav}.

However, every remote sensing-based monitoring effort depends on the accuracy and validity of the geographical boundaries of the monitored planting sites. Our location data integrity assessment considers multiple spatial indicators to identify potential discrepancies or inaccuracies in the georeferenced data provided. These indicators cover a comparative analysis of the planting sites and their surrounding environment, presence of other land cover within the reforestation site, and geometric characteristics of site boundaries, including nesting, intersection and administrative boundary alignments, and atypical geometric shapes. This approach will enable a comprehensive evaluation of spatial data quality that complements the ecological and socioeconomic success metrics currently used in restoration monitoring. Thus, the study's contributions are fourfold:

\begin{enumerate}
    \item A consolidated dataset of global reforestation efforts at the planting site-level.
    \item A location data integrity assessment of the georeferenced information based on a variety of indicators and data sources.
    \item  Sentinel-2 satellite imagery time series for most planting sites in a convenient format to facilitate further analysis.
    \item LLM-based text analysis of project description documents on contextual factors related to reforestation that are not well captured via remote sensing.
\end{enumerate}

Besides descriptive analytics, this dataset could also be used as a weakly-labelled training dataset for a diverse set of image recognition tasks, e.g.to monitor longevity of reforestation efforts. Weak labels in this context are image attributes that are usually derived from secondary data sources and therefore might be subject to substantial levels of noise. Despite this lack of precision, weak labels can still serve as valuable signals for model training.

\section*{Methods}

As illustrated in Figure 1, we compile publicly available information on reforestation efforts from different websites, harmonize the available data and augment them with planting site-level information on the presence of roads and human settlements, on tree and land cover change and on weather characteristics, among other things. In addition, we utilize freely available Sentinel-2 imagery (specifically Harmonized Sentinel-2 MSI: MultiSpectral Instrument, Level-1C) at 10 m resolution available from June 2015 onward to obtain relevant vegetation indices. 

In the following sections, we describe the dataset generation in more detail.
.

\subsection*{Site Data Acquisition}

To construct a global dataset of reforestation projects, we implement a systematic data collection process consisting of two main steps: 
\begin{itemize}
\item \textbf{Identification of reforestation efforts:} In a first step, we conduct extensive open Google searches to identify potential websites of interest using the following keywords: "reforestation", "carbon registry", "tree planting", "climate mitigation tree" and "carbon market". Table \ref{tab:list_websites} in the Appendix provides the list of websites selected to be of potential interest for our study.
\item \textbf{Data retrieval:} 
The majority of websites do not provide project information in an easily retrievable format. While some websites such as Tree Nation (\url{https://tree-nation.com}) maintain well-documented Application Programming Interfaces (APIs), many websites display information via dashboards that do not allow for direct downloads and thus require scraping the relevant information. Table \ref{tab:list_websites} provides details on the data collection mode and the reason why data collection was successful or not. It should be noted that a large share of websites of interest do not provide geographic information that allow pinpointing sites to a specific location. In most of these cases, the name of a larger administrative region is stated (e.g. "Yucatan, Mexico"), which, however, is insufficient to track reforestation progress via remote sensing. The retrievable location data are usually nested: A website hosts multiple reforestation projects which in turn are made up of multiple planting sites. We define a planting site as a closed, contiguous area with a non-zero area size. Consequently, if data are provided in complex polygonal structures, we break these up into individual polygons that adhere to our definition of a planting site. The data is stored at the site-level, where possible. During data retrieval, we capture metadata on project- and site-level, and on website-level.
\end{itemize}
\begin{figure}[ht]
    \centering
    \begin{subfigure}[b]{1\textwidth}
        \centering
        \includegraphics[width=1.0\textwidth]{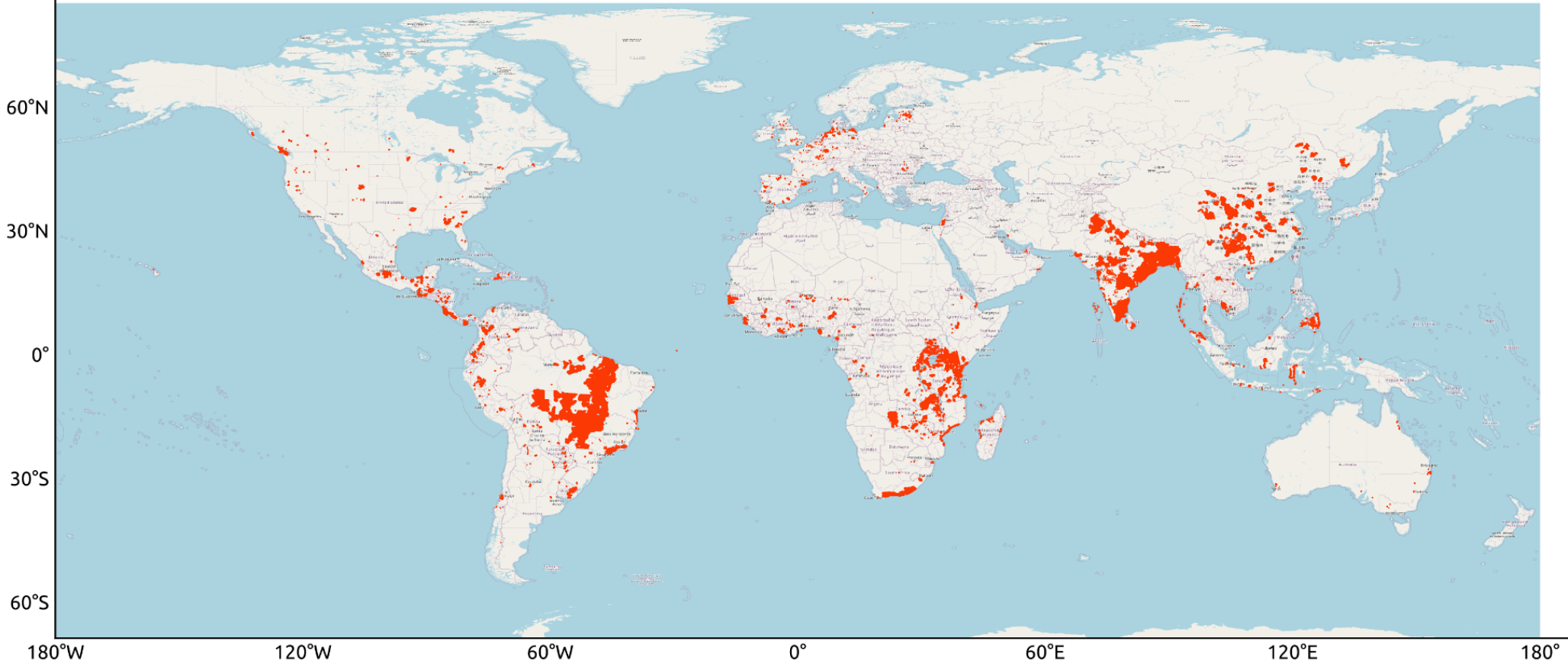}
        \caption{\delete{Reforestation sites, by location} \add{Reforestation site locations. Each red marker indicates a polygon of a reforestation site  area (as defined by our dataset: contiguous planting boundary or registered planted area).}}
        \label{fig:reforsites}
    \end{subfigure}
    \vfill
    \begin{subfigure}[b]{1.0\textwidth}
        \centering
        \includegraphics[height=4cm]{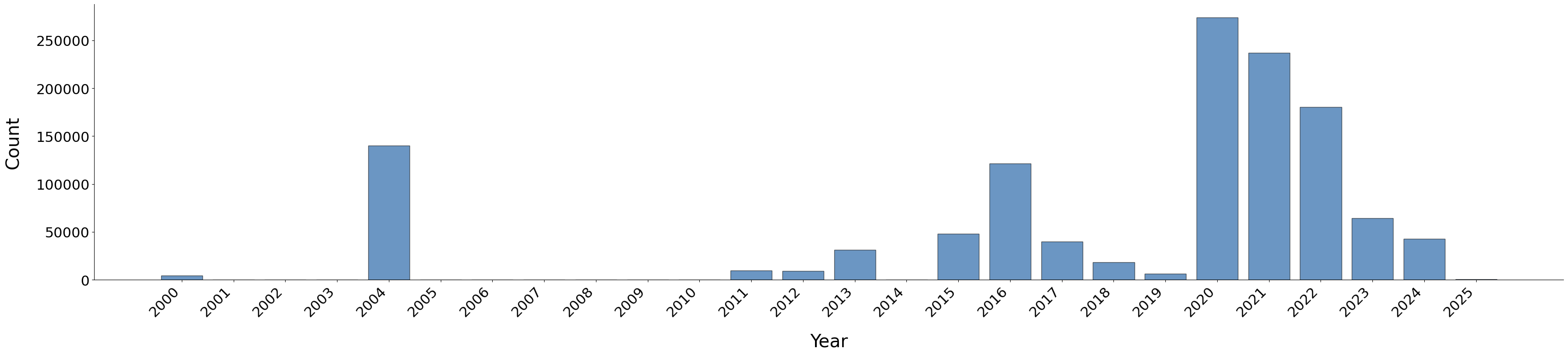} 
        \caption{\delete{Reforestation sites, by planting year}\add{Count of reforestation sites by planting year. Each bar shows the number of distinct sites planted in each year.}}
        \label{fig:count}
    \end{subfigure}
    \caption{\textbf{Overview of reforestation sites, by location and planting date.}}
    \label{fig:twosidebyside}
\end{figure}

\subsection*{Site Preprocessing}

As the data originates from different websites and is retrieved in different ways, data harmonization is needed, i.e. nomenclature and units of measurement need to be aligned. Furthermore, we assume a considerable amount of imperfect duplicates in the data. There are multiple reasons for that: First, since the VCM lacks vertical integration, multiple actors may advertise the same project at various layers of the market. For example, the project developer wants to sell carbon credits, and therefore showcases their work to inform about their offers. At the same time, the certification body is interested in positioning itself as a trustworthy major player in the certification market and, therefore, wants to display all projects it certified. Simultaneously, companies that buy carbon credits from that project to offset their emissions are inclined to use it for communications purposes, such as corporate social responsibility campaigns. Second, the data are partly crowdsourced with little quality checks. Third, nesting might occur where individual sites are combined and displayed as a larger site by another website. Fourth, not all projects provide actual site locations in the form of areas covered. In many cases, either names of larger administrative regions are stated or point locations are provided. Both make it difficult to track reforestation efforts as the areas of interest are not clearly delineated. Specifically, the following preprocessing steps have been pursued to address some of these issues:
\begin{itemize}
\item \textbf{Aligning nomenclature:} Aligning column names and units of measurement (e.g. hectares vs $\text{km}^2$) and other naming conventions within columns is a prerequisite for cross-source comparisons. Indicators with information as reported on the website are denoted with the suffix $\_reported$. Indicators estimated within this study are denoted with the suffix $\_derived$. It should be noted that one key information, namely the planting date, is subject to some uncertainty as organizations report dates that may differ in their definition: some provide the project registration date, some the intervention year, some the crediting starting period and some the actual planting dates. We summarize the different definitions in $planting\_date\_reported$ and give information about the definition type in the column $planting\_date\_type$.

\item \textbf{Filtering for afforestation/reforestation projects:}  
Although our initial site-acquisition screening targets reforestation initiatives, further filtering is necessary because some of our data sources also include unrelated activities. To keep the scope strictly to afforestation and reforestation, we exclude projects outside these categories. We implement this by using existing project classifications where available, and otherwise by applying keyword matches to project names and descriptions.

\item \textbf{Identifying special geometries:} Not all projects have polygonal site locations. However, clearly delineated areas are important for assessing and augmenting these sites. The same holds for site locations that are unrealistically large, e.g. spanning whole continents. Thus, we identify projects which closely resemble administrative areas (defined as overlapping each other with more than 98\% of their respective area) according to the GADM dataset  \cite{hijmans2018gadm} and store this information as the binary indicator $is\_exact\_administrative\_area$ in our dataset. We also calculate the circularity of each site stored as $circularity$. Geometries that closely resemble perfect circles (defined as a site being at least 98\% circular) as near-perfect circularity hints at buffered (point) locations that most likely do not reflect the actual extent of a planting site. This information is stored again as a binary indicator called $is\_perfectly\_circular$. For sites described by a single point geometry we define a 100 meters buffer around each point and store this new geometry in the column $geometry\_derived$. For polygon‐based sites, we create a 500 m outer buffer, where the ‘buffer’ specifically refers to the annular region between the original polygon boundary and its 500 m extension; this allows us to compare within‐polygon versus outside‐polygon signals for boundary accuracy and to identify any spatial leakage of greenness measurements.

\item \textbf{Identifying nested structures:} We note three different types of relationships between sites: subsets, duplicates, and supersets. If the area of the intersection of two sites constitutes more than 95\% of the two individual areas, respectively, we assume that they are duplicates of each other. If the intersection constitutes more than 95\% for one site, but less so for the other, we believe the former to be a subset of the latter and thus the latter to be a superset of the former. We, therefore, add the column $intersecting\_with$ to highlight the sites that intersect, and $nested\_in$ and $contains\_small\_polygon$ to highlight subsets and supersets. This approach is crucial for avoiding double-counting of restoration areas and understanding the hierarchical organization of efforts. As shown in Table 4, nested structures are most prevalent among smaller sites (e.g., the substantial proportion of nested areas in sites under 500 km²). In contrast, larger sites (above 1,000 km²) are rarely nested in other polygons, suggesting most large-scale projects are independent and not embedded within even larger areas. Recognizing these nested relationships is essential for accurate spatial analysis and for assessing the true extent and distribution of reforestation activities.
\end{itemize}

\subsection*{Data Augmentation}

To enhance the dataset’s analytical depth, we integrate remote sensing data and secondary environmental indicators to evaluate the quality and impact of reforestation efforts. These augmentation steps address various aspects such as tree cover change, infrastructure presence, climate conditions, land use transitions, and topographic features. By incorporating these additional layers of secondary data, we aim to facilitate the cross-validation of self-reported project information.

\subsubsection*{Project Descriptions}
 
Most sites or related projects provide a project description documenting project details. To extract species and planting‐date information from these descriptions, we employ two transformer‐based large language models:

\begin{itemize}
    \item \textbf{BERT–Q\&A:} We use the \texttt{bert-large-uncased-whole-word-masking-finetuned-squad} model \cite{DBLP:journals/corr/abs-1810-04805}, a large BERT model pre-trained with whole-word masking and fine-tuned on the squad benchmark, to extract the species planted.
    \item \textbf{DistilBERT–Q\&A:} We use the \texttt{distilbert-base-cased-distilled-squad} model \cite{sanh2019distilbert}, a lighter, faster distilled version of BERT fine-tuned on squad, to extract planting dates.
\end{itemize}

These choices are motivated by the ability of these models to derive responses strictly from the provided context, thus avoiding hallucinations (i.e., cases where the model generates information not supported by the input) and improving reproducibility. Given the projects' respective descriptions, we ask them the following questions:

\begin{itemize}
  
    \item What species were planted? Name each species mentioned.
   
    \item What is the planting date?
    \item What is the project start date?
    \item How many trees were planted?
\end{itemize}

Responses to each question are stored as separate indicators in the dataset. Some websites directly report the species planted, hence it is stored as the indicator $species\_planted\_reported$. In addition, we also extract information about the species planted from the project descriptions. We store this information as $species\_planted\_derived$. To ascertain the models' results, we manually checked 305 PDFs and compared them to the results provided by the models. The models correctly identified the species mentioned in 264 of the 305 PDFs and extracted accurate planting dates in 278 cases. When species were misidentified, the model typically returned general terms such as “native species,” “indigenous tree species,” “native vegetation types,” or “locally adapted species.” Likewise, when planting dates were incorrect, the model often produced broad timeframes (e.g., “minimum 40 years,” “4–6 years”) or general targets (e.g., “by 2030”).

\subsubsection*{Tree Cover Loss}\label{sec:tree_cov_change}

We also consider tree cover loss to be an important indicator for monitoring reforestation efforts, particularly with respect to their additionality and permanence. Have the sites experienced measurable tree cover loss prior to planting? Have the sites experienced tree cover loss in the years following planting? 

In our study, we assess tree cover loss patterns using the Global Forest Watch 2023 v1 dataset\cite{hansen2013high}. This dataset detects forest loss events at a minimum mapping unit of 30 m resolution per pixel. Tree cover loss is defined as a stand-replacement disturbance or complete removal of the tree canopy. Also it does not distinguish between permanent deforestation and temporary losses (e.g., due to fire or logging followed by re-growth). 
We use this dataset to estimate the proportion of tree cover loss within each site for the five years before planting, the year immediately preceding planting, and the five years after planting, where applicable, for the period 2000 to 2023.

\subsubsection*{Infrastructure Presence}
The extent of human presence within reforestation sites may provide information on the accuracy of the sites' reported geographical boundaries: if the supposed planting site overlaps with a human settlement, then the site's boundaries are unlikely to be accurate. Consequently, we report the proportion of area under human settlements within each site using the GHS-BUILT-S dataset from the P2023 release of the Copernicus' Global Human Settlement Layer (GHSL)\cite{pesaresi2023ghs}. 
Furthermore, we use NASA's gROADSv1\cite{center2010global} dataset to estimate the length of roads (in km) per square kilometer within a given site. 

\subsubsection*{Climate Features}
Climatic conditions such as precipitation and temperature have a strong influence on forest growth and reforestation success \cite{oogathoo2024precipitation, ma2023precipitation}. Using the historic WorldClim API \cite{harris2020version}, we extract the climate features, i.e. average monthly rainfall and minimum and maximum temperature experienced within each reforestation site at different points in time: at planting date and one year, two years and five years after planting using the site midpoint geometries.
\subsubsection*{Site Terrain}
Furthermore, we include the elevation and slope variables derived from the Shuttle Radar Topography Mission (SRTM) \cite{farr2007shuttle}\delete{also based on reforestation site midpoint geometries. Terrain features are critical for understanding the physical landscape and terrain of a reforestation site, which directly influences reforestation efforts' success or failure } For each site, we extract all intersecting $90 m \times 90 m$ raster cells to compute mean elevation and mean slope indicators. However, we note that SRTM’s 90 m resolution may not capture very fine-scale topographic conditions.

\subsubsection*{Land Cover and Land Use}
Land use conflicts are prevalent and a major driver of deforestation\cite{gibbs2010tropical}. Identifying areas where reforestation would directly compete with existing livelihoods can help prevent potential conflicts with local communities that depend on these resources, which could lead to the failure of these reforestation projects. Therefore, understanding the prevalence of different land cover types at project sites is a potentially relevant indicator for reforestation success. To unveil land cover change patterns within the project area at various points in time, we use the Global 2000-2020 land cover and land use change (GLAD) dataset\cite{potapov2022global} to extract proportions of reforestation site area with following land cover transitions: cropland from tree, cropland to tree, short vegetation after tree loss, stable cropland 2000-2020 and permanent water.
While the GLAD-derived Global Forest Watch layer provides annual, wall-to-wall coverage, its classification accuracy varies by biome and transition type. For instance, the tropical-forest gain class exhibits a false-negative rate of approximately 52\%, meaning over half of real gain events in tropical regions may go undetected. Overall, global false-positive rates for forest loss and gain are roughly 13\% and 23.6\%, with corresponding false-negative rates of 12.2 \% and 26.1 \% \cite{hansen2013high}. These limitations imply that our estimates of both loss and gain could under- or over-represent true changes, and thus should be interpreted with appropriate caution.

\subsubsection*{Annual Tree Cover Change Indicators}\label{sec:treecov_change}
Existing datasets such as \cite{hansen2013high} and \cite{potapov2022global} do not provide tree cover extent and gain estimates on an annual basis. The Normalized Difference Vegetation Index (NDVI) \cite{tucker1979red} is widely used to monitor reforestation and vegetation changes. \cite{pansit2024detecting, kirbizhekova2023monitoring} have shown its effectiveness in assessing reforestation success and tracking changes in vegetation cover over time. According to \cite{huang2021analysis}, the NDVI is positively correlated with standing stock, particularly on larger spatial scales. However, \cite{pansit2024detecting} indicates that NDVI alone may not always capture substantial changes in forest cover, suggesting the need for integrated monitoring approaches. According to \cite{kirbizhekova2023monitoring}, combining NDVI with other indices, such as the Normalized Difference Red Edge (NDRE) index, can provide a more comprehensive assessment of afforestation and reforestation dynamics. Thus, as vegetation indices of choice, we select the NDVI as a measure for presence or absence of tree cover, the NDRE for measuring a vegetation's health and the Soil-Adjusted Vegetation Index (SAVI) to correct the NDVI from any soil reflectance when the vegetation cover is sparse (e.g. when trees are still young and canopy cover sparse). All these indices are calculated from Sentinel-2 imagery COPERNICUS/S2\_HARMONIZED" as we aim to cover most of our reforestation sites from their earliest dates. Table \ref{tab:vegetation_overview} gives a summary of the definitions of the three indices used.

\begin{table}[ht]
\centering

\renewcommand{\arraystretch}{1.5} 
\begin{tabular}{>{\arraybackslash}m{4cm} >{\arraybackslash}m{6cm} >{\arraybackslash}m{4.5cm}}
\toprule
\textbf{Index} & \textbf{Definition/Description} & \textbf{Formula} \\ 
\midrule
\textbf{NDVI (Normalized Difference Vegetation Index)} & Provides information about the presence or absence of green vegetation, and can indicate vegetation health in some cases. &
\(\frac{\text{Near Infrared (NIR)} - \text{Red}}{\text{Near Infrared (NIR)} + \text{Red}}\) \\
\textbf{NDRE} & Identifying subtle stress factors and chlorophyll variations in trees& 
\(\frac{\text{Near Infrared (NIR)} - \text{RedEdge}}{\text{Near Infrared (NIR)} + \text{RedEdge}}\) \\
\textbf{SAVI} & To correct NDVI for the influence of soil brightness in areas where vegetative cover is low.& 
\(\frac{\text{NIR} - \text{RedEdge}}{\text{NIR} + \text{RedEdge}+0.5}\)  * (1+0.5)\\
\bottomrule
\end{tabular}
\caption{\textbf{Overview of NDVI, NDRE and SAVI definitions.} Using bands from Sentinel-2.}
\label{tab:vegetation_overview}
\end{table}

To limit the impact of seasonal fluctuations on our vegetation indices, we calculate annual vegetation indices based on the top three greenest months as measured by monthly NDVI averages in 2023 (in Africa: Jan, Apr, May, in Asia: Jun, Jul, Aug, in Europe: Apr, May, Jun, in South America: Feb, Mar, Oct). As illustrated in Figure 3, the NDVI changes by years around planting, but behaves similarly to the NDRE index and the SAVI.\\

\begin{figure}[ht]
    \centering
    \begin{subfigure}[b]{0.49\textwidth}
        \centering
        \includegraphics[width=\textwidth]{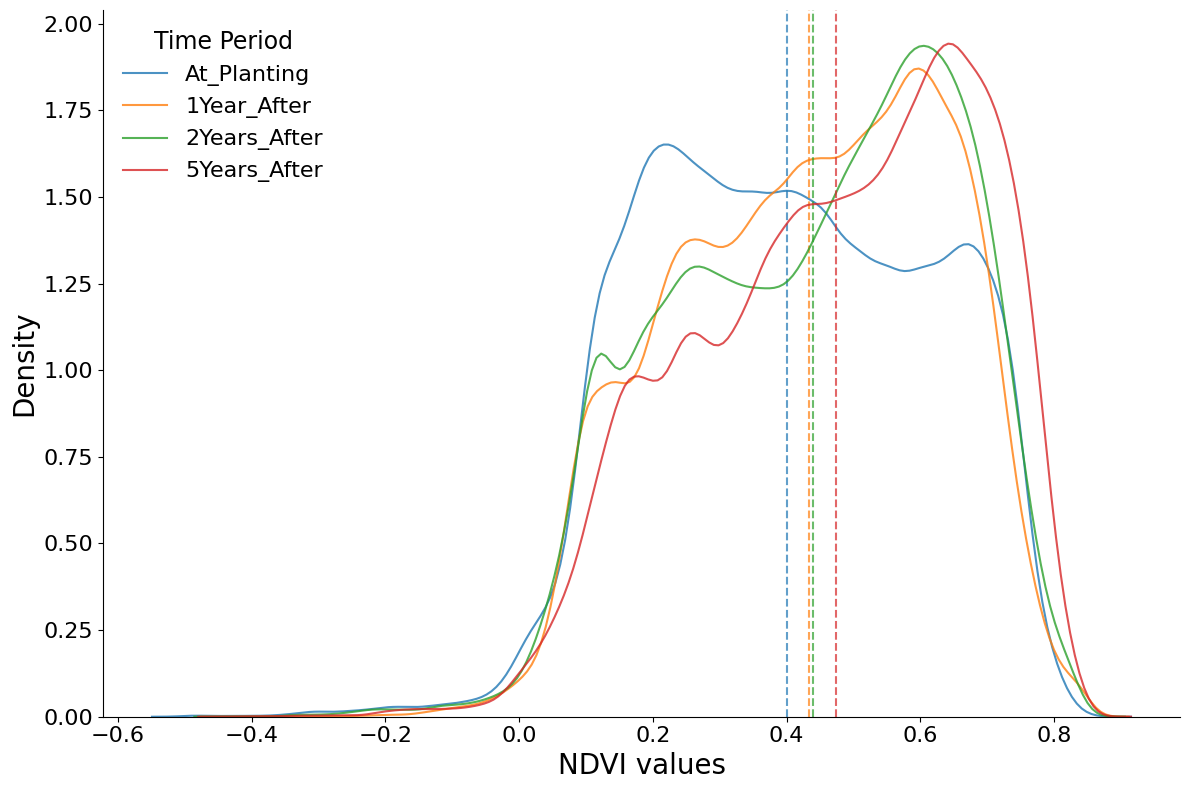}
        \caption{NDVI distribution}
        \label{fig:ndvi_dist}
    \end{subfigure}
    \begin{subfigure}[b]{0.49\textwidth}
        \centering
        \includegraphics[width=\textwidth]{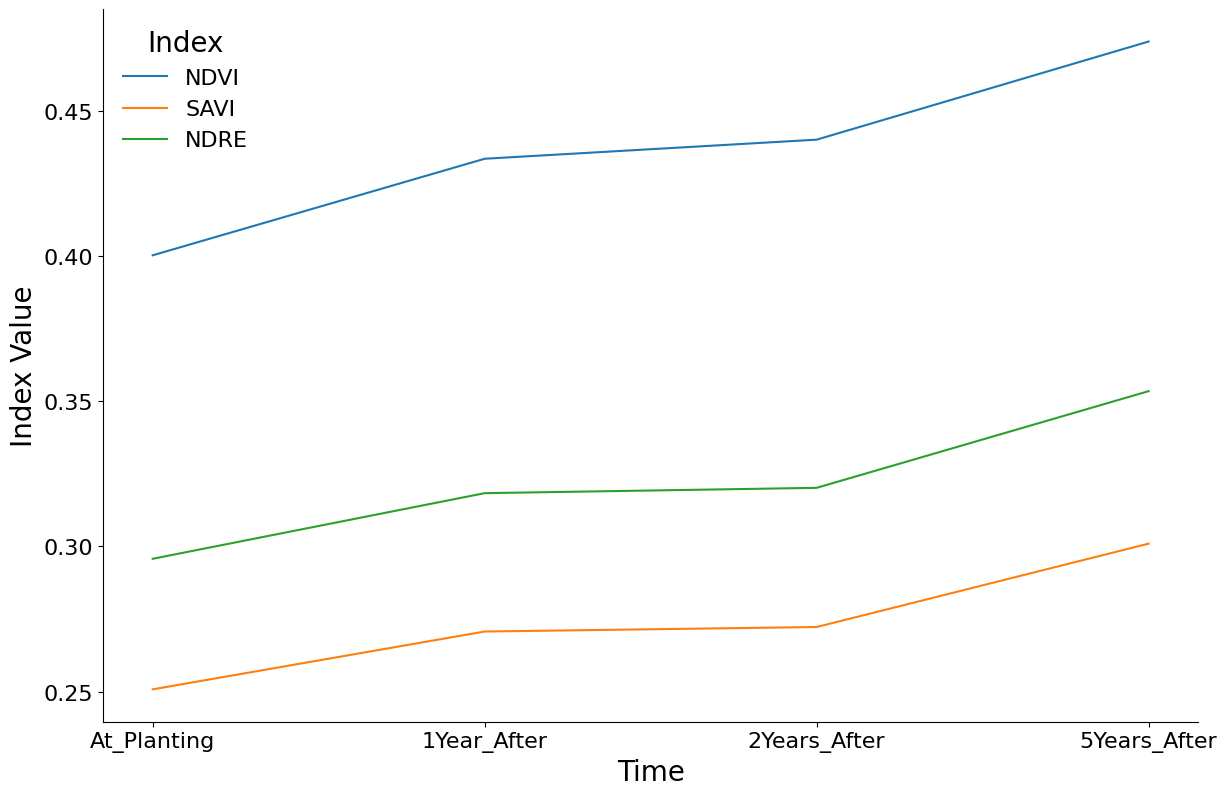}
        \caption{Reforestation Change Using Indices}
        \label{fig:indices}
    \end{subfigure}
    \caption{\textbf{Distribution of vegetation indices.}}
    \label{fig:ndvi_res}
\end{figure}

The NDVI distribution shifts progressively to the right—from a mean of 0.39 at the (likely) planting date to 0.47 five years later, indicating a clear increase in vegetation greenness. The narrowing of the distribution and the higher peak density at year 5 suggest not only recovery but also greater uniformity in canopy cover. Interestingly, a subset of sites already exhibited NDVI values above 0.60 at planting, highlighting that some ``reforestation” areas in our database were dense forest from the outset. This overall trend confirms NDVI as a plausible and sensitive proxy for monitoring post‑planting.
However, we also consider the NDVI, NDRE and SAVI as our main indicators of reforestation success over time as local variations between the indicators may apply.

\subsection*{Location Data Integrity Assessment}\label{sec:ldis}

The accuracy of the provided delineations of the planting sites is a crucial element for correctly assessing and quantifying reforestation efforts from space. 
However, even planting sites with perfectly delineated polygons do not guarantee successful revegetation without proper planting protocols, periodic monitoring, and attention to surrounding land‐use activities.
We try to understand a site's geographic integrity by triangulating different perspectives: 

\begin{enumerate}
\item We check for other human infrastructure or water bodies within the planting area. If built-up areas, roads, or water bodies cover more than 10\% of the planting site, we assume that the delineation of the actual planting area lacks accuracy. 
\item  We consider land cover classifications and land cover conversions after planting. If other land cover classes are present and cover 20\%  or more of the planting area, we take this as an indication that the planting is not accurately delineated.
\item We account for possible double counting in projects by checking the intersecting ($intersecting\_with$) and nested polygons ($nested\_in$). If the planting areas are nested within or intersecting with each other, we take this as an indication of inaccurate delineation and/or potential double counting.
 
\item We check whether the site boundaries resemble (subnational) administrative areas available through GADM \cite{hijmans2018gadm}. Named $exact\_admin\_area$, this binary indicator takes a value of $1$ if a site shares more than 98\% of its area with an administrative area and vice versa, $0$ otherwise.
\item We assess if the provided site boundaries are perfectly circular, hinting at inaccurate actual planting site boundaries, e.g. by approximating them via buffered point locations. Again stored as a binary indicator, $polygon\_circle\_oval$ takes the value of $1$ if the site boundaries are 95\% circular, $0$ otherwise. 
\item The invalid geometries indicator $project\_geometries\_invalid$ is used to identify whether the reported geometries are valid. Specifically, this check ensures that the provided geometry is a properly closed polygon, which is necessary for consistent spatial analysis and monitoring.
\item  We check for the presence of forest on the planting date. If the forest at the planting covers 20\% of the planting site area or more, we take this as an indicator of the planting site not accurately delineated.
\item We check for the presence of stable cropland area within the planting site between the years 2000 and 2020. If stable cropland covers 20\% or more of the planting area, we take this as an indication that the planting is not accurately delineated
\end{enumerate}

\begin{table}[htbp]
\centering
\caption{\textbf{Location Data Integrity Score (LDIS) indicators.} Data completeness indicates the percentage of non-missing entries for each indicator. All indicators use a binary (0/1) value range and equal weight (1).}
\label{tab:integrity_indicators}
\begin{adjustbox}{width=\textwidth}
\begin{tabular}{|l|l|p{4cm}|c|}
\hline
\textbf{Indicator Name} & \textbf{Database Field Name} &\textbf{Definition}&\textbf{Data Completeness (\%)} \\
\hline
road\_presence & total\_road\_length\_km & Area within the site covered by roads &82 \\
\hline
built\_area\_presence & built\_area\_2018 &Area within the site covered by buildings or constructions & 83 \\
\hline
forest\_at\_planting\_glad & treecover\_atplanting & Area within the site covered by trees at planting time&99 \\
\hline
other\_landcover\_score & other\_land\_cover\_area\_2020& Area within the site covered by other landcover e.g shrubs,grassland& 98 \\
\hline
nesting\_polygon & nested\_in & Checking if a site is contained inside another site fully&100 \\
\hline
intersecting\_Polygon & intersecting\_with & checking if part of the site is overllaping with another site & 100 \\
\hline
exact\_admin\_area & exact\_admin\_area & Checking if the site is  an exact representation of an administration area as defined by the GADM data &100 \\
\hline
perfect\_circle\_indicator & polygon\_circle\_oval & Checking if the site is almost a perfect circle &100 \\
\hline
geometry\_validity & project\_geometries\_invalid & Checking for  provided geometries if they represent a closed boundaries geometry or point geometry &100 \\
\hline
stable\_cropland\_score & stable\_cropland\_cover\_area\_2020& Area covered by stable cropland from 2000 to 2020 within the site& 87 \\
\hline
\end{tabular}
\end{adjustbox}
\end{table}

Most of the indicators require an area to be calculated. Since some planting sites are originally just provided as point locations, we use their derived geometry (i.e. buffering a site's point location with a 100m buffer) for further analysis. However, this naive approach most likely does not capture the true extent of a planting site. Thus, we expect the LDI scores for this group of sites to be lower on average than for planting sites that come with a delineated area. Figure 4 shows the LDIS distribution for planting sites in the dataset for those two groups separately.

\begin{figure}[ht]
    \centering
    \begin{subfigure}[b]{0.49\textwidth}
        \centering
        \includegraphics[width=\textwidth, keepaspectratio]{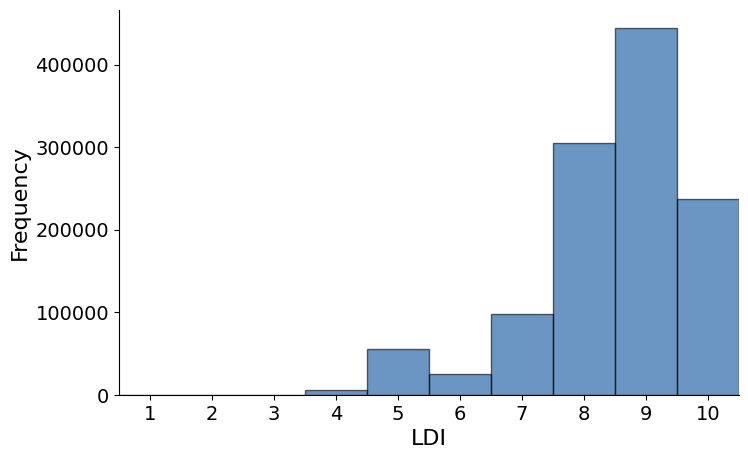}
        \caption{Sites with provided area geometries}
        \label{fig:ldis_withoutpoint}
    \end{subfigure}
    \begin{subfigure}[b]{0.49\textwidth}
        \centering
        \includegraphics[width=\textwidth, keepaspectratio]{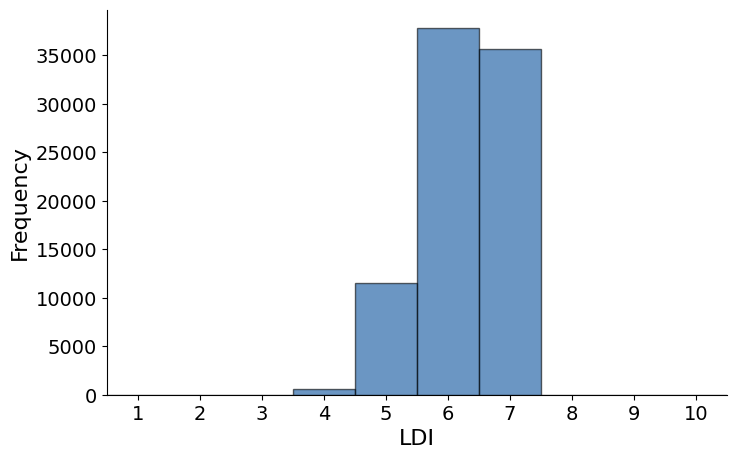}
        \caption{Sites with buffered point geometries}
        \label{fig:ldis_withpoint}
    \end{subfigure}
    \caption{\textbf{Location data integrity score (LDIS) distribution, by geometry type.}}
    \label{fig:ldis_by_geometry}
\end{figure}
For the group of reforestation sites with area geometries (Figure 4a), 79\% of these sites have an LDIS of 9 and less, implying that only 21\% sites meet all our quality indicators. For the group of sites with buffered point geometries (Figure 4b), the number of projects that score perfectly on the LDI assessment reduces to 0\%.

\section*{Data Records}
The dataset associated with our study can be found in a public data repository:\url{https://dataverse.harvard.edu/previewurl.xhtml?token=edd23c6e-bd89-4052-932c-2039e11b9d17}. It contains georeferenced information on reforestation efforts around the globe, including hierarchically-organized metadata. Information are stored on the level of individual reforestation sites, where possible. If site-specific data is not available -- for example, when geographic information are not provided as actual planting sites, but via higher-level location names -- information are stored at the project- or host-level. 
Site data is further augmented with indicators on tree cover change, forest loss, land cover change, and the presence of human settlements and infrastructure within these reforestation sites, among others. A detailed description of each indicator in our dataset can be found in Tables \ref{tab:metadata_derived},\ref{tab:secondary_data} and LDI indicators extracted from the variables in \ref{tab:integrity_indicators}. The dataset spans approximately 33 years and is based on the analysis of nearly 50 organizations working in the reforestation space. Links to the sources are provided both in Table \ref{tab:list_websites} as well as in the column \textit{url} of the dataset. Primary datasets from these websites are available upon request or retrieval can be repeated using the code published alongside the dataset. All the data records are consolidated into a single geographic parquet file (.parquet-format). 

Alongside the georeferenced dataset, we download Sentinel-2 median composite scenes for each planting site, even those lacking pre-computed cloud flags, and for each scene, compute the within polygon cloud fraction using the QA60 band. We then retain only those composites where cloud cover over the site footprint is less than 20 \%.
These images correspond to four distinct time points: the planting year, one year before planting and one, two, and five years after planting. For each of these periods, we selected the composite image from the month with the highest Normalized Difference Vegetation Index (NDVI). We used these composite images to calculate NDVI, SAVI, and NDRE, as analyzed and discussed in section ``Annual Tree Cover Change Indicators" above, by focusing on those months with top NDVI values.

\section*{Technical Validation}%

\phantomsection
\addcontentsline{toc}{section}{Technical Validation}
\label{sec:val}

The purpose of this dataset is to consolidate publicly accessible information on reforestation efforts, augment these with indicators derived from metadata and secondary data and assess the integrity of the location data provided for these efforts on the level of individual planting sites. The dataset sets out to be comprehensive in scope and instead of excluding sites from the beginning, we aim to flag data quality constraints through a well-defined selection of indicators. To validate the scope of projects included in the dataset, we compare it with the Voluntary Registry Offsets (VRO) Database of the Berkeley Carbon Trading Project (\url{https://gspp.berkeley.edu/research-and-impact/centers/cepp/projects/berkeley-carbon-trading-project/offsets-database}), which contains an updated list all carbon offset projects from the registries of the four major independent carbon credit programmes (namely Verra, Gold Standard, Climate Action Reserve and ACR). These four programmes account for the vast majority of the global VCM \cite{carbon_offset_guide_programs}. We find that we capture all projects that are also in the VRO database, which make up 85\% of all the area covered in our dataset (see Table \ref{tab:projdist}). \add{Table \ref{tab:projdist} shows how carbon-credit projects are distributed across the world’s leading registries. “Projects” is the number of distinct carbon-credit projects registered under each host. “Sites” is the total number of individual planting polygons reported by those projects (if a host does not publish polygon data, its entry is “–”). “Georeferenced” counts only the sites with valid latitude and longitude coordinates. “Area” is the sum of all georeferenced site polygons expressed in square kilometres, and “Avg. Area per Site” is the mean polygon size for those georeferenced sites. Together, these metrics reveal both the scale (total area) and granularity (number of sites) of each registry’s portfolio.} As both our database as well as the VRO database store the original unique project identifiers of the respective carbon credit programmes (in our case $project\_id\_reported$), the databases can easily be matched for further analysis.

\begin{table}[ht]
\centering
\caption[Project distribution across hosts]{\textbf{ Project distribution across hosts.} Number of reforestation projects and planting sites under major carbon-credit programmes.  

\textit{Column definitions:}  
\textbf{Projects} = total number of reforestation projects;  
\textbf{Sites} = total number of discrete planting locations (– indicates data not available);  
\textbf{Georeferenced} = number of sites with valid spatial coordinates;  
\textbf{Area (km$^2$)} = cumulative georeferenced planting area in square kilometers;  
\textbf{Avg. Area per Site (km$^2$)} = mean area per georeferenced site.}
\label{tab:projdist}
\begin{adjustbox}{width=\textwidth}
\begin{tabular}{|l|r|r|r|r|r|}
\hline
\textbf{Project Host}          & \textbf{Projects} & \textbf{Sites} & \textbf{Georeferenced} & \textbf{Area (km$^2$)} & \textbf{Avg. Area per Site (km$^2$)} \\
\hline
Verra                         & 636               & 1,225,618      & 1,225,618              & $3.52\times10^{7}$     & 28.71                              \\
\hline
Gold Standard                 & 71                & –              & –                      & –                       & –                                  \\
\hline
Climate Action Reserve (CAR)  & 513               & 513            & 513                    & 1,683                   & 3.28                               \\
\hline
American Carbon Registry (ACR)& 283               & –              & –                      & –                       & –                                  \\
\hline
Other Hosts                   & 44,316            & 69,648         & 63,668                 & 6,579,569               & 103.34                             \\
\hline
\end{tabular}
\end{adjustbox}
\end{table}

Validating how many of all reforestation projects in the world we do not capture, proves more challenging. During data collection, we observed that publicly accessible information on government-led projects is especially sparse. As around 70\% of global forestland is under "legal and administrative authority" of governments \cite{pokorny2019forests}, we assume that government-led reforestation efforts are under-represented in the current dataset.

On the level of planting sites, we observe that the 1\% largest sites make up about 90\% of all the total reforested area in our database (see Table \ref{tab:site_summary_stats}).Table~\ref{tab:site_summary_stats} summarizes the distribution of reforestation site areas. “Size Bin” refers to the range of individual site areas (in km\textsuperscript{2}). “Count” is the number of sites that fall into each size bin, and “Count (\%)” shows what percentage of all sites that represents. “Total Area” is the cumulative area of all sites in each bin, and “Total (\%)” reflects the share of the total reforested area. “Nested Area” refers to overlapping areas within larger planting polygons, and “Nested (\%)” is the proportion of bin area that is nested.

\begin{table}[ht]
\centering

\captionsetup{font=small,labelfont=bf}
\caption[Site summary statistics]{\textbf{Distribution of reforestation site areas.} Summary of site-area bins and their contributions.  

\textit{Column definitions:}  
\textbf{Size Bin (km$^2$)} = range of site area;  
\textbf{Count} = number of sites within the bin;  
\textbf{Count (\%)} = percentage of total sites;  
\textbf{Total Area (km$^2$)} = sum of areas of all sites in the bin;  
\textbf{Total (\%)} = percentage of total area;  
\textbf{Nested Area (km$^2$)} = area of of sites fully inside other sites within the bin;  
\textbf{Nested (\%)} = percentage of nested area relative to bin's total area.}
\label{tab:site_summary_stats}
\begin{adjustbox}{width=\textwidth}
\begin{tabular}{|l|r|r|r|r|r|r|}
\hline
\textbf{Size Bin (km$^2$)} & \textbf{Count} & \textbf{Count (\%)} & \textbf{Total Area (km$^2$)} & \textbf{Total (\%)} & \textbf{Nested Area (km$^2$)} & \textbf{Nested (\%)} \\
\hline
< 10        & 1,225,267 & 99.68 & 53,188    & 1.17  & 6,858   & 12.89 \\
\hline
10--50      & 1,423     & 0.12  & 28,885    & 0.63  & 4,302   & 14.89 \\
\hline
50--100     & 232       & 0.02  & 16,500    & 0.36  & 2,799   & 16.96 \\
\hline
100--500    & 636       & 0.05  & 156,765   & 3.44  & 24,145  & 15.40 \\
\hline
500--1000   & 254       & 0.02  & 183,050   & 4.01  & 1,908   & 1.04  \\
\hline
1000--2000  & 637       & 0.05  & 851,289   & 18.67 & 0       & 0.00  \\
\hline
2000--5000  & 488       & 0.04  & 1,622,356 & 35.58 & 0       & 0.00  \\
\hline
> 5000      & 237       & 0.02  & 1,648,053 & 36.14 & 0       & 0.00  \\
\hline
\end{tabular}
\end{adjustbox}
\end{table}

By disaggregating the LDI score by area size, we observe that larger sites score considerably lower than smaller sites (see Figure 5).

\begin{figure}[ht]
    \centering
    \begin{subfigure}[b]{0.49\textwidth}
        \centering
        \includegraphics[width=\textwidth, keepaspectratio]{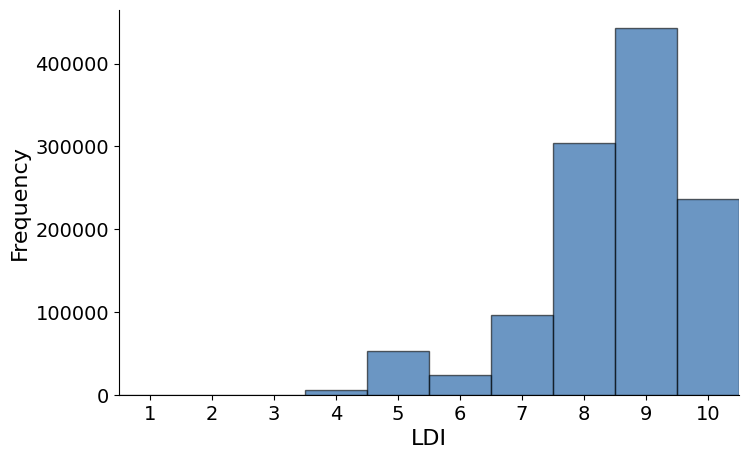}
        \caption{Sites $< 5\text{km}^2$}
        \label{fig:ldis_by_area_small}
    \end{subfigure}
    \begin{subfigure}[b]{0.49\textwidth}
        \centering
        \includegraphics[width=\textwidth, keepaspectratio]{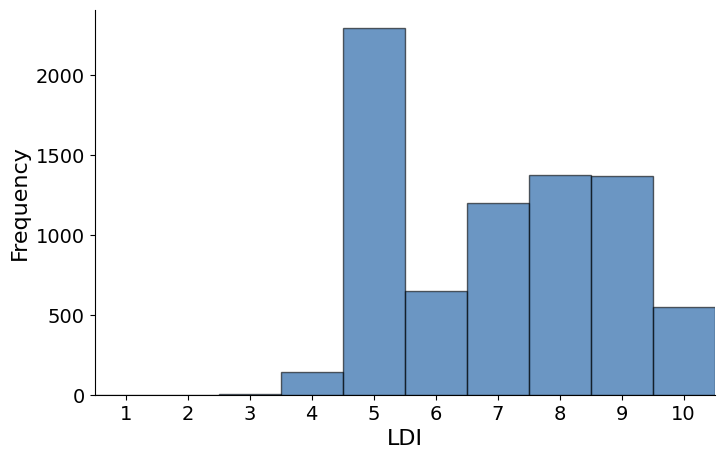}
        \caption{Sites $\geq 5\text{km}^2$}
        \label{fig:ldis_by_area_large}
    \end{subfigure}
    \caption{\textbf{Location data integrity score (LDIS) distribution, by area size.}}
    \label{fig:ldis_by_area}
\end{figure}

Looking at the underlying LDIS indicators for the largest sites, we further observe that most of these sites closely resemble an administrative area or are almost perfectly circular. In those cases, the likelihood that infrastructure or other land cover areas are also covered by a given site is quite high, thus explaining the low LDI scores. Figure 6 provides examples of what the sites for a given LDIS look like.

\begin{figure}[ht]
    \centering
    \begin{subfigure}[b]{0.32\textwidth}
        \centering
        \includegraphics[width=\textwidth, keepaspectratio]{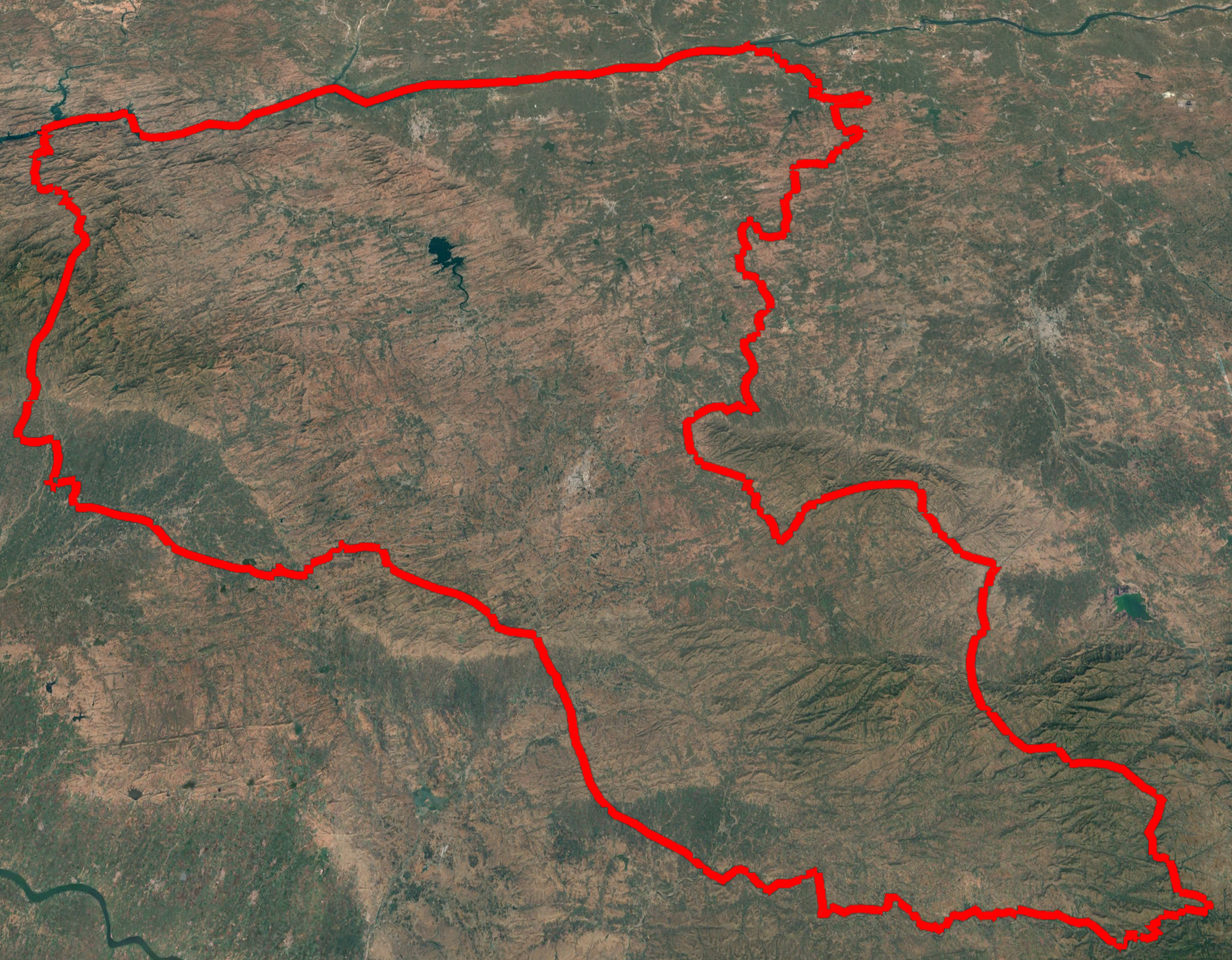}
        \caption{LDIS 2, with area $5438.7\text{km}^2$}
        \label{fig:sat_img_ldis2}
    \end{subfigure}
    \hfill
    \begin{subfigure}[b]{0.32\textwidth}
        \centering
        \includegraphics[width=\textwidth, keepaspectratio]{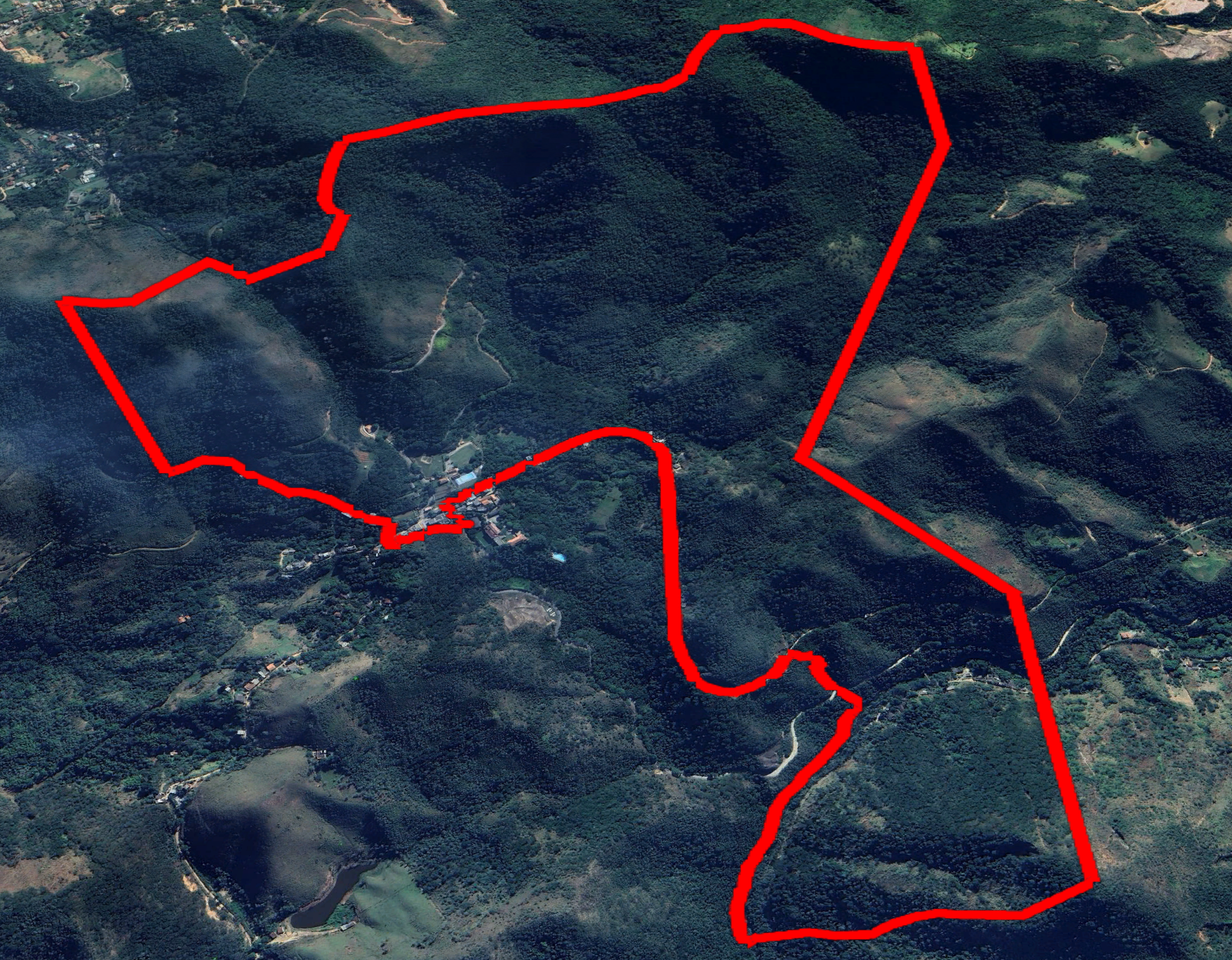}
        \caption{LDIS 6, with area $2.62 \text{km}^2$ }
        \label{fig:sat_img_ldis6}
    \end{subfigure}
    \hfill
    \begin{subfigure}[b]{0.32\textwidth}
        \centering
        \includegraphics[width=\textwidth, keepaspectratio]{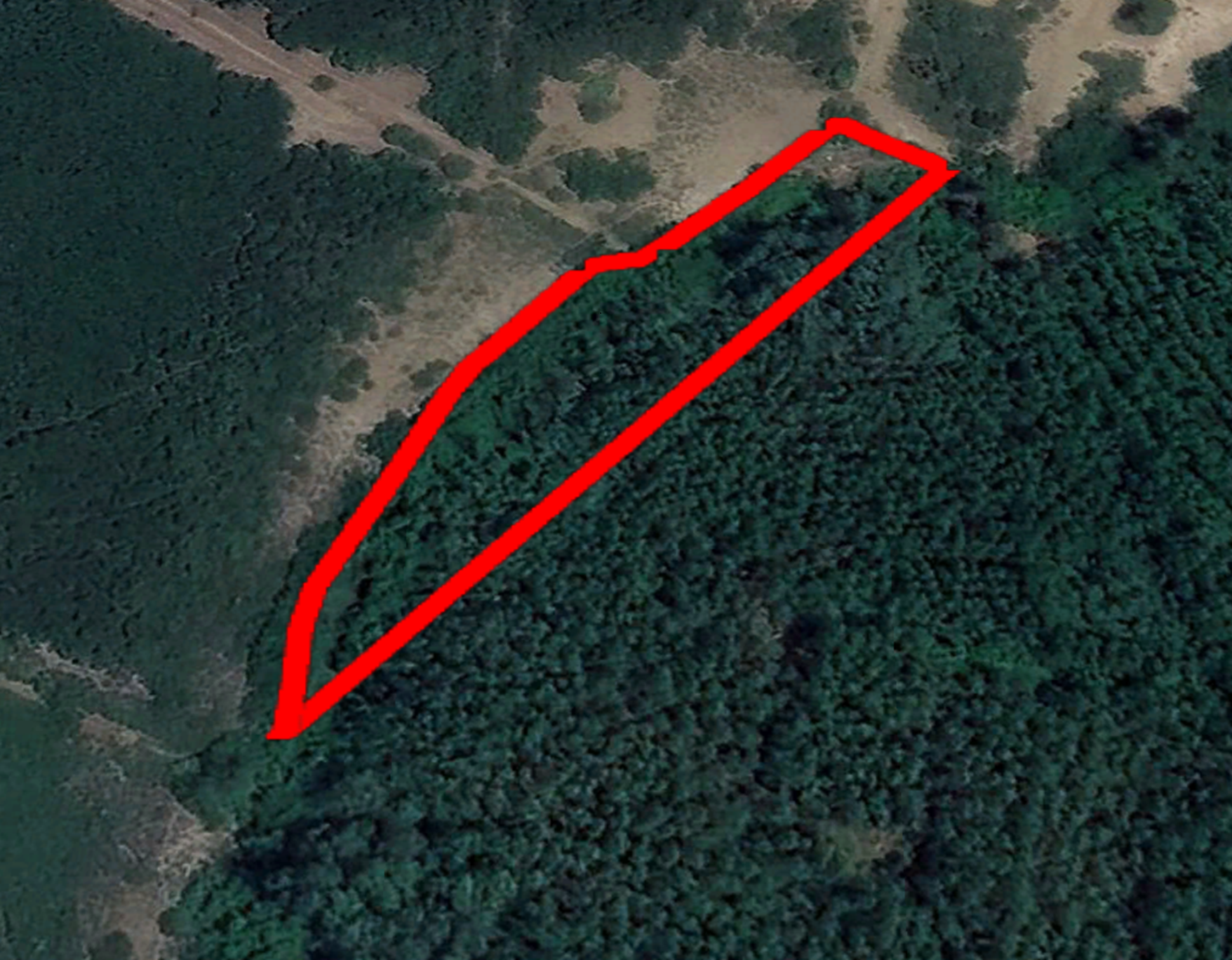}  
        \caption{LDIS 9, with area $0.00311 \text{km}^2$}  
        \label{fig:sat_img_ldis9}
    \end{subfigure}
    \caption{\textbf{Exemplary sites, by location data integrity score (LDIS).}}
    \label{fig:sat_img_ldis}
\end{figure}

For a subset of 250 manually annotated satellite images from 250 randomly selected reforestation sites, we capture the binary variable $Geometry\_ Capturing\_Planting\_Area$, which is \textit{yes}, if the reforestation area is accurately and comprehensively captured without other activities visible on-site (see Table \ref{tab:quality_groups}).

\begin{table}[htbp]
\centering
\caption{\textbf{Comparing location data integrity score (LDIS) against manually annnotated images.}}
\label{tab:quality_groups}
\begin{tabular}{lrrr}
\toprule
Geometry Capturing Planting Area & No & Yes & Total \\
LDI Score &  &  &  \\
\midrule
6 and below & 87.5\% & 12.5\% & 8 \\
7-8 & 10.2\% & 89.8\% & 128 \\
9 & 4.4\% & 95.6\% & 114 \\
Total & 25 & 225 & 250 \\
\bottomrule
\end{tabular}
\end{table}

At LDIS of 9, for 95.6\% of the sites there is agreement with the manually annotated cases which is an increase from the 12.5\% and 89.8\% for an LDIS of 6 and below and between 7 and 8, respectively.
To further understand the quality of the planting site boundaries exactness and any leakages, we set a 500 m buffer around each planting site polygon provided. Figure 7 shows the mean NDVI change per period across planting sites and their respective buffer areas, with a 95\% bootstrap confidence interval around it.

\begin{figure}[ht]
    \centering  
    \begin{subfigure}[b]{0.49\textwidth}
        \centering
        \includegraphics[width=\textwidth, keepaspectratio]{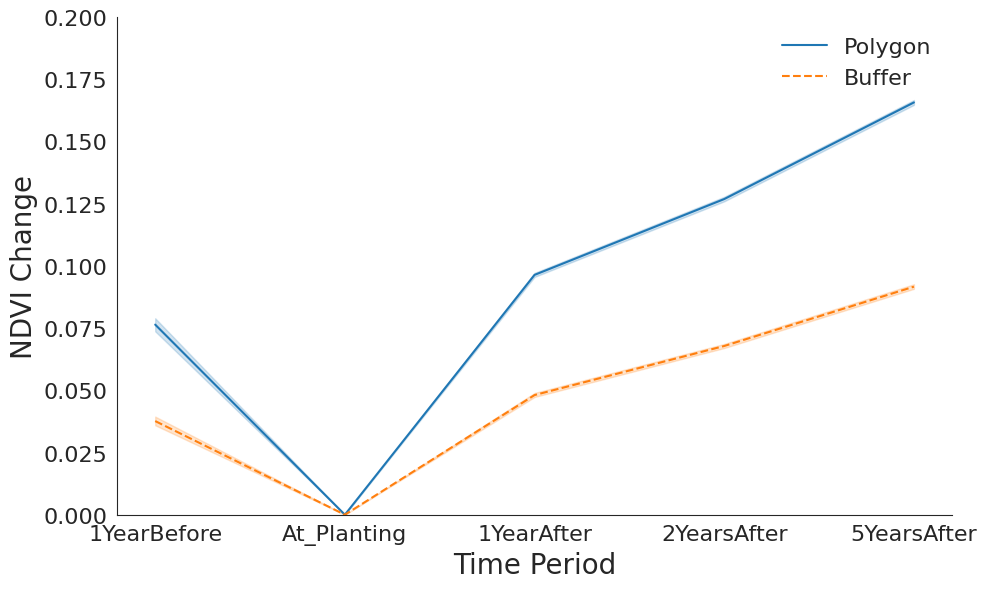}
        \caption{Sites $< 5\text{km}^2$}
        \label{fig:all}
    \end{subfigure}
    \hfill
    \begin{subfigure}[b]{0.49\textwidth}
        \centering
        \includegraphics[width=\textwidth, keepaspectratio]{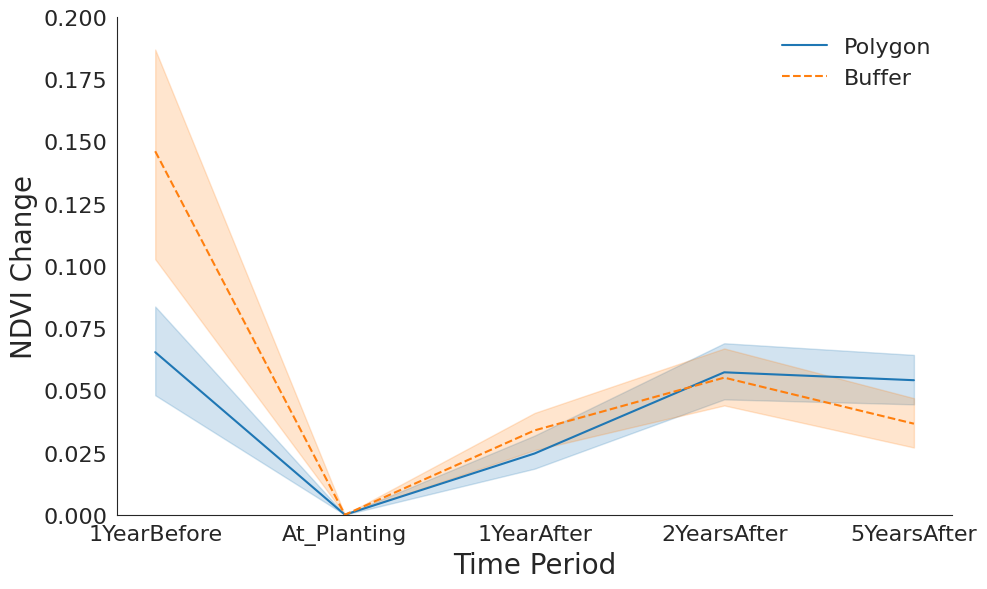}
        \caption{Sites $\geq 5\text{km}^2$}
        \label{fig:allbx}
    \end{subfigure}  
  
    \caption{\textbf{NDVI change over time.} Comparing planting sites and their buffer area, by area size. Uncertainty is represented by bootstrapped 95\% confidence intervals.}
    \label{fig:buffer_res_expanded}
\end{figure}

Although the NDVI changes of the buffer and planting sites in our data set are notably different, both show a similar increase over time. Potential reasons for that are spillover and edge effects and growth of shrubs and possible presence of mature trees around the site the buffer. Similar cases have been witnessed in other studies such as \cite{simkins2022haiti} and \cite{mahato2023comparative} monitoring the buffer and planting using vegetation indices. 
 
Interestingly, we also observe considerably higher NDVI values in the year pre-planting compared to the planting year. Potential reasons for this are multiple such as clearing sites from shrubs, grass and weeds before planting or reforestation efforts being a consequence of previous wildfires or other adverse events. Manually inspecting satellite imagery for a random subset of planting sites that show significant NDVI drop from the pre-planting to the planting year does not reveal obvious common causes for this phenomenon as it ranges from shrubs and grass land before planting to other cases having forest before planting. However, we consider this phenomenon (including land management activities like tillage and land preparations before planting \cite{mulla2013twenty, pettorelli2007early}) to be relevant especially in the context of additionality and permanence of reforestation efforts and therefore regard further investigations in that direction as an excellent use case of our database. Moreover, in Figure 7a, the CI around the mean NDVI change line is very narrow, indicating more precise estimates at smaller sites. By contrast, Figure 7b shows wider shaded bands for larger polygons, reflecting greater variability in mean NDVI change in Larger polygons in our database.

To investigate the significance of the planting effect on the NDVI more rigorously, we apply a Difference-in-Difference (DiD) approach. In the DiD design, we consider planting as treatment with NDVI values before and after treatment within and outside the planting area (defined by a 500 m buffer around the planting site) as the control, respectively, as our outcome of interest. We subset our dataset to sites that have NDVI observations available for every pre-, at and post-planting period (excluding 5-years before planting due to data sparsity), thereby ensuring a balanced panel.While we recognize that land-cover histories may differ — with planting areas often being previously deforested or degraded, and buffers potentially retaining existing vegetation or representing recovering secondary forests — we assume the common trend assumption (i.e. vegetation growth within and around the planting area would be similar if trees had not been planted) to hold. This assumption is supported by the preceding analysis (see Figure7, which shows that NDVI change trends between planting areas and their buffers are closely aligned not only prior to planting but also in the years immediately following it.

The DiD approach is implemented as a linear regression with a time-treatment interaction effect (see Equation \ref{eq:diff_in_diff}).
The general form of the Difference-in-Differences (DiD) model is:
\begin{equation}\label{eq:diff_in_diff}
    Y_{it} = \text{intercept} + \beta_1 \cdot \text{g}_i + \beta_2 \cdot \text{t}_j + \beta_3 \cdot (\text{g}_i \times \text{t}_j) + \epsilon_{it},
\end{equation}

\noindent where $Y_{it}$ represents the outcome of interest, namely the average NDVI at the planting site $i$ at time $t$.    $\text{g}_i$  is a binary variable that indicates whether it is a treatment (planting site) or a control ( buffer), $\text{t}_j$ indicates time as  either before or after the planting date, and $\epsilon_{it}$ is the error term. The interaction term $\text{g}_i \times \text{t}_j$ describes the effect of interest, for which $\beta_3$ captures the effect of treatment (the DiD estimate). The results of the DiD approach are shown in Table \ref{tab:did_regression}.

\begin{figure}[ht]
    \centering
        \includegraphics[width=\textwidth, keepaspectratio]{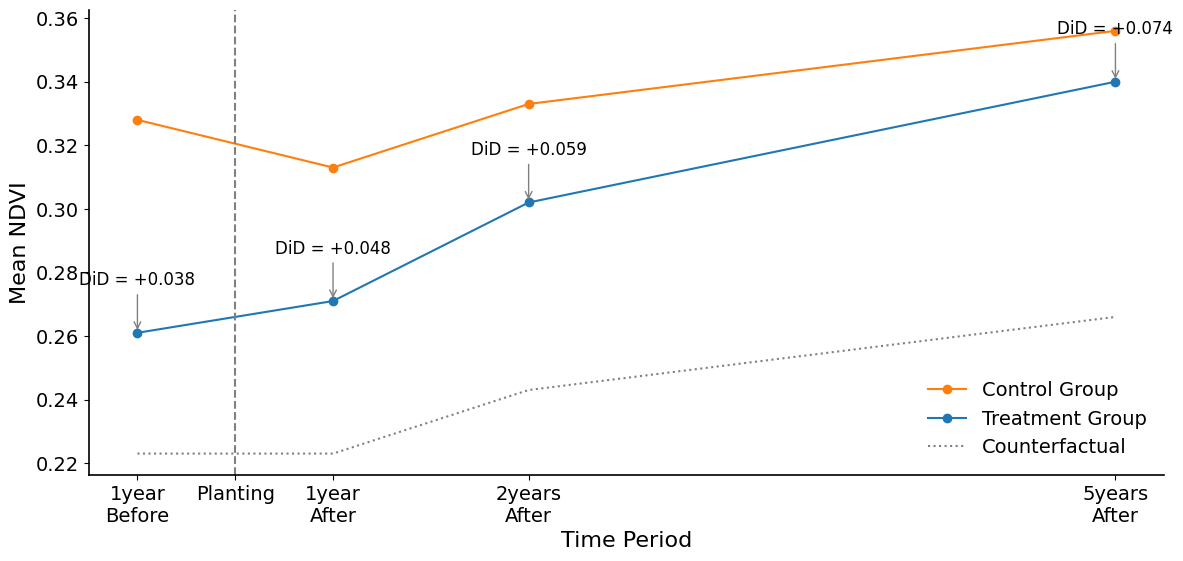}
        \label{fig:ke_line}
         \caption{\textbf{Evolution of the NDVI over time.} Subset of sites with NDVI data for all relevant dates around planting. NDVI averaged across sites. The control group is composed of buffer areas (\add{defined by a 500 m buffer around the planting site}) around respective planting sites. DiD values are estimated against planting date NDVI values and correspond to the difference in NDVI between the Treatment group and the Counterfactual. The counterfactual are model-based NDVI predictions excluding the interaction (\textit{DiD}) component.}
    \label{fig:DID_PLOT}
\end{figure}

\begin{table}[!htbp]
\centering
\caption{\textbf{Difference‐in‐Differences regression results for NDVI.}}
\label{tab:did_regression}
\begin{adjustbox}{width=\textwidth}
\begin{tabular}{@{\extracolsep{5pt}}lcccc}
\hline
\hline \\[-1.8ex]
& \multicolumn{4}{c}{\textit{Dependent variable: NDVI}} \\
\cline{2-5}
\\[-1.8ex]
& \multicolumn{1}{c}{One Year Before}
& \multicolumn{1}{c}{One Year After}
& \multicolumn{1}{c}{Two Years After}
& \multicolumn{1}{c}{Five Years After} \\
\\[-1.8ex]
& (1) & (2) & (3) & (4) \\
\hline \\[-1.8ex]
\textbf{ Intercept} & 0.290$^{***}$ & 0.265$^{***}$ & 0.265$^{***}$ & 0.265$^{***}$ \\
& (0.001) & (0.000) & (0.001) & (0.001) \\
\textbf{treat (g)} & -0.105$^{***}$ & -0.090$^{***}$ & -0.090$^{***}$ & -0.090$^{***}$ \\
& (0.002) & (0.001) & (0.001) & (0.001) \\

 \textbf{time (t)} & 0.038$^{***}$ & 0.048$^{***}$ & 0.068$^{***}$ & 0.091$^{***}$ \\
& (0.002) & (0.001) & (0.001) & (0.001) \\
\textbf{treat*time (gt)} & 0.038$^{***}$ & 0.048$^{***}$ & 0.059$^{***}$ & 0.074$^{***}$ \\
& (0.002) & (0.001) & (0.001) & (0.001) \\
\hline \\[-1.8ex]
 \textbf{Observations} & 139250 & 469500 & 469500 & 469500 \\
  \textbf{\( R^2 \)} & 0.061 & 0.081 & 0.097 & 0.132 \\
 \textbf{Adjusted \( R^2 \)} & 0.061 & 0.081 & 0.097 & 0.132 \\
 \textbf{Residual Std. Error} & 0.208 (df=139246) & 0.170 (df=469496) & 0.181 (df=469496) & 0.184 (df=469496) \\
\textbf{F Statistic} & 2995.626$^{***}$ (df=3; 139246) & 13736.170$^{***}$ (df=3; 469496) & 16738.387$^{***}$ (df=3; 469496) & 23880.764$^{***}$ (df=3; 469496) \\
\hline
\hline \\[-1.8ex]
\multicolumn{1}{l}{\textit{Note:}}
& \multicolumn{4}{r}{$^{*}$p $<$ 0.1; $^{**}$p $<$ 0.05; $^{***}$p $<$ 0.01}
\end{tabular}
\end{adjustbox}
\end{table}

The DiD regression results in Table \ref{tab:did_regression} confirm a statistically significant positive impact of reforestation as measured by the NDVI one year before and one, two and five years after planting. Interestingly, the NDVI values of buffer areas are higher than in their respective planting sites with the difference growing around as presented by the negative coefficients of the treatment term $g$ in regression (1) and (2), thus hinting at site clearing efforts at planting sites. This is supported by the positive time term $t$ before planting. After planting, it takes more than two years of tree growth to compensate for the different offset of buffer and planting sites as shown by the difference between the treatment term $g$ and the planting effect $gt$. The positive time term $t$ after planting implies that no extensive logging or clearing has taken place since planting.

To validate that these temporal changes especially in the buffer areas are site-specific trends and not driven by higher-level dynamics such as climate change, we apply the synthetic control method to identify the "counterfactual" trend. Although we perform this validation step exemplary for one country, namely Kenya, we assume that the results are by-and-large representative for other regions in the world. To create a synthetic control group to our planting sites (in this context we consider "planting" as the treatment), we randomly select 10,000 points in Kenya, calculate their NDVI at planting and allocate each of the values to a bucket derived from NDVI distribution of the reforestation sites. We then use weights from these NDVI distributions to calculate the weighted mean across the buckets, resulting in NDVI averages for the synthetic control group. Figure 9 compares the NDVI changes over time for the planting sites, their respective buffer areas and their synthetic controls for sites located in Kenya.

\begin{figure}
        \includegraphics[width=\textwidth, keepaspectratio]{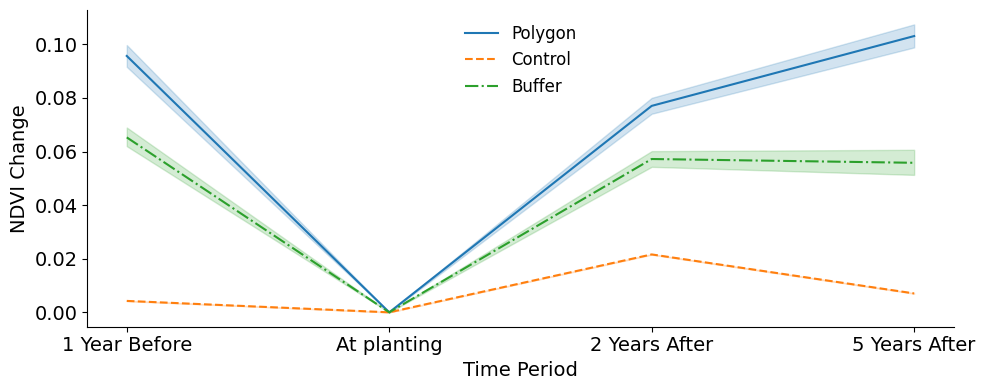}
        \label{fig:ke_line}
         \caption{\textbf{NDVI change over time in Kenya.} Comparing planting sites, their buffer areas and corresponding synthetic control groups  all sites  have $< 5\text{km}^2$ .}
    \label{fig:kecombined_results}
\end{figure}

Similar to the NDVI changes over time presented in Figure 9 for all sites, the Kenyan sites show similar trends for the planting sites and their buffer areas. However, the synthetic control group remains largely flat as one would expect in the absence of common dynamics. This validates our initial assumptions that a) noticeable site-specific planting effects exist, and b) the sites are to a large extent not properly delineated or other mechanisms cause the buffer area to behave similar to the corresponding planting site.

In a final step, we validate our indicators derived from secondary data by the randomly sampled 250 sites with annotated historic high-resolution images of these sites accessed via Google Earth Pro and Global Forest Watch (GFW) Map before, at and after planting. We check for the presence of roads, buildings, and forests of each reforestation site in a binary manner. We compare our manual annotations with indicators using a confusion matrix and calculate accuracy and F1 score.

\begin{figure}[ht]
    \centering
    \begin{subfigure}[b]{0.32\textwidth}
        \centering
        \includegraphics[width=\textwidth, keepaspectratio]{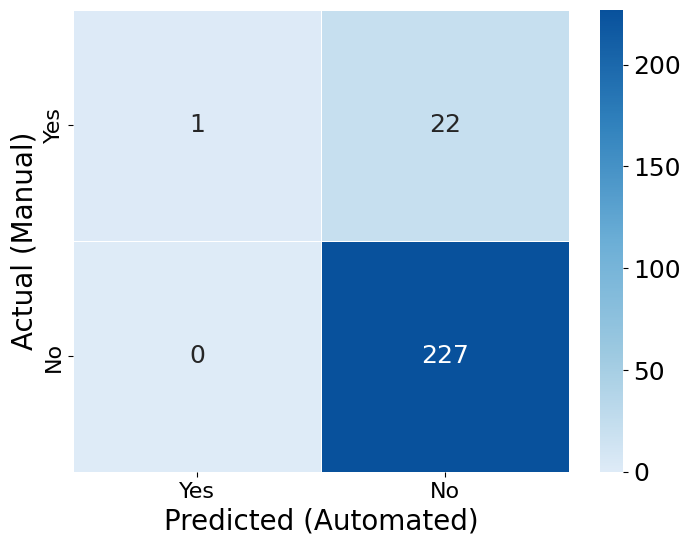}
        \caption{Presence of roads}
        \label{fig:road}
    \end{subfigure}
    \hfill
    \begin{subfigure}[b]{0.32\textwidth}
        \centering
        \includegraphics[width=\textwidth, keepaspectratio]{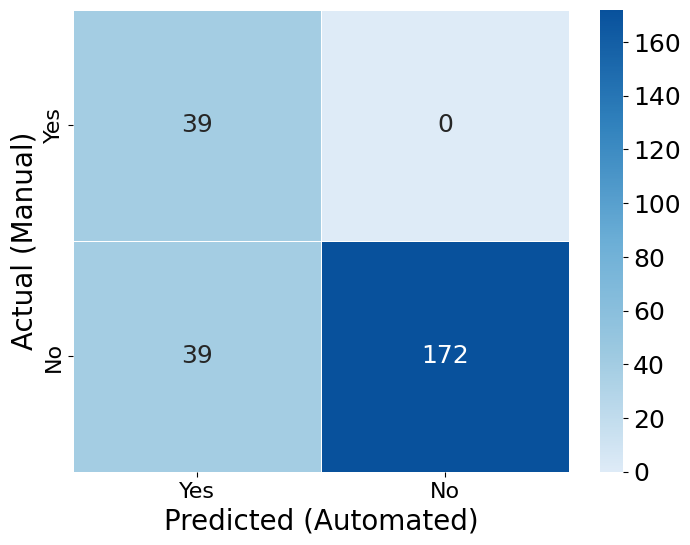}
        \caption{Presence of built-areas}
        \label{fig:built}
    \end{subfigure}
    \hfill
    \begin{subfigure}[b]{0.32\textwidth}
        \centering
        \includegraphics[width=\textwidth, keepaspectratio]{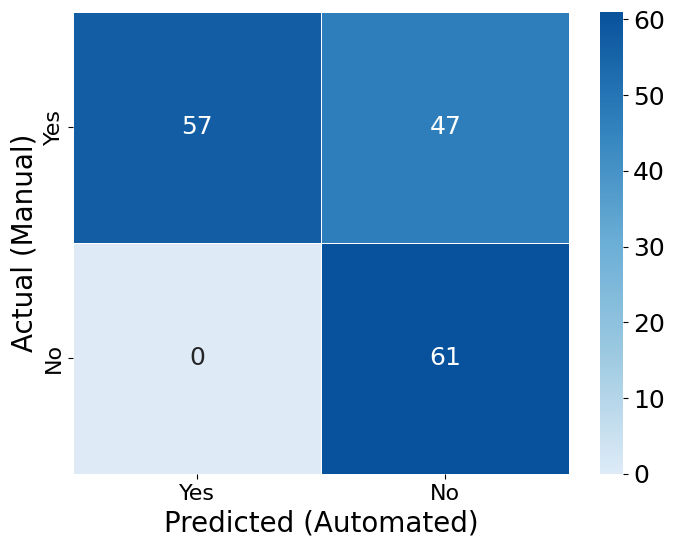}  
        \caption{Forest at planting}  
        \label{fig:glad}
    \end{subfigure}
    \caption{\textbf{Comparing derived indicators with manually-labelled annotations.}}
    \label{fig:conf}
\end{figure}

For the presence of roads within the sampled planting sites, we calculate an accuracy of 91\% precision and an F1 score of 0.87, effectively identifying roads compared to manual annotations, with minor discrepancies in recognizing smaller roads. For the presence of built areas, we calculate an accuracy of 84\% and an F1 Score of 0.86, indicating good precision in mapping built areas. Differences arise due to varying definitions of built-up areas, as some open spaces and roads are classified as built areas in the automatic framework using the GHS BUILT settlement data. For the presence of forests at the time of planting, we compare our indicators to manual annotations using high-resolution images from Google Earth Pro and GFW, as well as lower-resolution Sentinel-2 and Planetscope images for cases where there are no timely high-resolution images. For detecting the presence of forests, we calculate an accuracy of 71.51\% (F1 Score: 0.71), thus performing moderately well. A potential explanation with the comparatively low alignment of manual annotations and dataset indicators can be found in the varying definitions of what classifies as a forest.

\section*{Usage Notes}

.

This dataset sets out to provide perspectives on reforestation success of reforestation efforts around the globe. Specifically, it provides an assessment of the integrity of the location data provided for each reforestation site -- an essential pillar for any downstream remote monitoring effort. However, the impact of these reforestation efforts is multi-dimensional. The dataset barely considers ecological, legal, managerial and socio-economic perspectives and consequently cannot be used to assess these dimensions and/or the overall quality of a given reforestation project.

Although we made efforts for this dataset to be as comprehensive as possible, there are multiple reasons why some reforestation efforts are not included: First, reforestation efforts without any publicly available information on the internet are excluded. While this might hold true for only a tiny fraction of projects that aim to participate in the VCM, we got the impression that especially government-led planting initiatives are much less well documented publicly. Second, since the data collection is based on keyword-specific Google searches, we might miss out on some initiatives that do not align with the chosen keywords. While we are confident that this is a minor issue for projects that host information in English, we might miss out more substantial parts of the reforestation space that provides information only in languages other than English. However, since the VCM, especially the carbon credit certification part, is dominated by organizations registered in the US and Switzerland, we expect that most projects that plan to participate in the VCM will have information in English listed. Third, some websites mention active reforestation efforts, but provide too little detail or data in machine-readable format (e.g. providing maps of site locations without annotations in .pdf-format) to add it to the dataset. Since we expect this to apply to a non-negligible share of projects, we flag those websites in Table \ref{tab:list_websites} accordingly. Further, in order to use the dataset effectively, consider the following:

\begin{itemize}
    \item Software for analysis: The dataset is compatible with various geographic information system (GIS) software, including QGIS, ArcGIS, and Google Earth Engine. It can also be used with most geospatial processing libraries of popular programming languages such as Python and R. 
    \item Metadata interpretation: Both geographic data and project descriptions in the dataset are taken \textit{as is} from the respective websites. We do not correct any of these information, but use secondary data to assess their validity. Refer to Tables \ref{tab:integrity_indicators}, \ref{tab:metadata_derived} and \ref{tab:secondary_data} for a detailed description of each indicator and its source.
    \item Hierarchical data: The dataset provides information on the level of planting sites. Each planting site has its own identifier assigned (\textit{site\_id\_reported}). Most sites are part of a reforestation project. A reforestation project usually consists of multiple planting sites and has its own identifier (\textit{project\_id\_reported}). While some reforestation projects may stretch across multiple countries, they are usually confined within the national boundaries of one country, denoted by its ISO-3 country code (\textit{iso3}). In addition, most websites host more than one reforestation project. Websites are listed with their \textit{url} and the website's name (\textit{host\_name}). The hierarchical structure of the data and the cross-level relationships need to be taken into account when conducting any analysis on the dataset.
    \item Planting date: As already noted in Section, there is considerable uncertainty related to the indicator $planting\_date\_reported$ as the underlying definitions planting date may vary across hosts. Please consider the indicator $planting\_date\_type$ and potential definitions from the hosts' websites for more details.
    \item Temporal analysis: The dataset enables temporal analysis of reforestation sites. However, although we use the same data sources across years and pay attention to flag methodological changes that may limit comparability over time, undeclared methodological changes may still occur. Users can track changes over time by comparing tree cover and vegetation-related indices (e.g. NDVI) at, one year after, two years after and five years after the planting date, if applicable. Tree loss and land cover conversion indicators are also available for the years prior to planting.
    \item Integration with other datasets: Users may integrate this dataset with other environmental datasets for a more comprehensive analysis of reforestation's impact on biodiversity, land use, and climate change. This can be mainly done via the geographic location of the planting sites (\textit{geometry\_reported}). Alternatively, since we report the original identifiers both for the sites and the projects, if available, they, too, can be used to integrate site- and project-specific information.
    \item Caution with interpretation: While the dataset provides a standardized assessment of the location data integrity, users should exercise caution when drawing conclusions about the overall reforestation success as reforestation success can be influenced by various factors not captured by the dataset as mentioned before. 
    .
    \item Data retrieval and re-use: The dataset is intended for scientific use and for supporting monitoring and verification processes in the carbon markets. While some programmes such as the ESA Copernicus Sentinel 2 (\url{https://developers.google.com/earth-engine/datasets/catalog/COPERNICUS_S2_HARMONIZED})explicitly allow for reuse, the terms and conditions of use may vary across websites.
\end{itemize}

\subsection*{Uncertainty}
The dataset relies on reported data and a variety of highly processed secondary data sources that both come with their own intricate error frameworks. It is beyond the scope of the paper to quantify and integrate the various sources of uncertainty that are inherent to this dataset. Just to name a few: Geographic boundaries of planting sites are prone to data entry errors and likely depend on the accuracy of the GPS device. In NASA's gROADS dataset, the underlying road representation dates range from the 1980s to 2010 and originate from different national sources with varying data qualities. Global land cover datasets exhibit considerable uncertainties \cite{cui2023comparison, venter2022global}. These limitations have to be especially considered for any future uses that rely on the correct approximation of the underlying error structure, e.g. for hypothesis testing. 
\subsection*{Figure legends}

This section contains our figures legends and titles.

\textbf{Figure 1  The reforestation data generation workflow.} We follow these steps in our reforestation data collection.

\textbf{Figure 2  Overview of reforestation “sites” in our dataset, by geographical location and planting date.} 
\textbf{a} Reforestation site locations.  Each red marker indicates the polygon boundary of a reforestation site (defined in our dataset as either a contiguous planting area or registered planted zone).
\textbf{b} Count of reforestation sites by planting year.Each bar represents the number of distinct sites planted in a given year.

\textbf{Figure 3  Distribution of vegetation indices.} 
\textbf{a}, NDVI distribution. The normalized difference vegetation index (NDVI) values across reforestation sites, showing the variation in vegetation density for periods atplanting,1 year after,2 years after and 5 years after planting.  \textbf{b}, Reforestation change using indices. Visualization of reforestation progress derived from vegetation index changes : Soil Adjusted Vegetation Index(SAVI),NDVI and Normalized Difference Red Edge index(NDRE) over time.

\textbf{Figure 4 Location data integrity score (LDIS) distribution, by geometry type.}  
\textbf{a}, Sites with provided area geometries. Shows the LDIS distribution for reforestation sites where detailed polygon geometries were available.  
\textbf{b}, Sites with buffered point geometries. Displays the LDIS distribution for sites where only point locations were provided and area was inferred via buffering.

\textbf{Figure 5 Location data integrity score (LDIS) distribution, by area size.}  
\textbf{a}, Sites smaller than 5~km\textsuperscript{2}. Shows the LDIS distribution for reforestation sites with a reported or estimated area under 5~km\textsuperscript{2}.  
\textbf{b}, Sites equal to or larger than 5~km\textsuperscript{2}. Displays the LDIS distribution for larger reforestation sites.

\textbf{Figure 6  Exemplary sites, by location data integrity score (LDIS).}  
\textbf{a}, Site with LDIS 2, area 5,438.7~km\textsuperscript{2}.  
\textbf{b}, Site with LDIS 6, area 2.62~km\textsuperscript{2}.  
\textbf{c}, Site with LDIS 9, area 0.00311~km\textsuperscript{2}.

\textbf{Figure 7 NDVI change over time.} Comparing planting sites and their buffer areas, by area size. Uncertainty is represented by bootstrapped 95\% confidence intervals.  
\textbf{a}, Sites smaller than 5km\textsuperscript{2}.  
\textbf{b}, Sites equal to or larger than 5km\textsuperscript{2}.

\textbf{Figure 8  Evolution of the NDVI over time.} Subset of sites with NDVI data for all relevant dates around planting. NDVI averaged across sites. The control group is composed of buffer areas ( (defined by a 500 m buffer around the planting site)around respective planting sites. DiD values are estimated against planting date NDVI values and correspond to the difference in NDVI between the Treatment group and the Counterfactual. The counterfactual are model-based NDVI predictions excluding the interaction (\textit{DiD}) component.

\textbf{Figure 9  NDVI change over time in Kenya.} Comparison of planting sites, their buffer areas, and corresponding synthetic control groups. All sites are smaller than 5km\textsuperscript{2}.

\textbf{Figure 10  Comparing derived indicators with manually labelled annotations.}  
\textbf{a}, Presence of roads.  
\textbf{b}, Presence of built-up areas.  
\textbf{c}, Forest at planting.

\section*{Code availability}

The programs used to generate the dataset are Python 3.10 and Google Earth Engine (GEE). The code and instructions to reproduce the dataset are available on 
\url{https://github.com/Societal-Computing/Forest_Monitoring}.

\section*{Acknowledgements}

The work is supported by funding from the Alexander von Humboldt Foundation and the Federal Ministry of Education and Research (Bundesministerium für Bildung und Forschung) of Germany.

\section*{Author contributions statement}

A.J., S.A., T.K., A.T. and I.W. conceived the presented idea and developed the methodology. A.J., S.A. and T.K. acquired the data. A.J. performed the computations. A.J., S.A., T.K., A.T. and I.W. analyzed the results. A.J., S.A. and T.K. wrote the manuscript with support of  A.T. and I.W. helped supervise the project. All authors reviewed and edited the manuscript.

\section*{Competing interests}

The authors declare no competing interests.

\section*{Figures \& Tables}

\begin{table}[ht]
\small
\centering

\begin{tabular}{|l|c|c|p{2cm}|}
\hline
Name & Status & Reason & Collection mode\\
\hline
8 Billion Trees \href{https://8billiontrees.com/planting-projects/}{\faExternalLink} & Unsuccessful& Location Available, Exact Site Geometry Unavailable&  \\
Afr100 \href{https://afr100.org/}{\faExternalLink} & Unsuccessful& Location Available, Exact Site Geometry Unavailable&\\
All4Trees \href{https://projects.all4trees.org}{\faExternalLink} & Unsuccessful& Location Available, Exact Site Geometry Unavailable&\\
African Development Bank \href{https://mapafrica.afdb.org/en/projects}{\faExternalLink} & Unsuccessful& Could not computationally retrieve data&\\
Americanforests \href{https://www.americanforests.org/}{\faExternalLink} & Successful& Point Geometry Available& API\\
Anthesis \href{https://www.climateneutralgroup.com/en/climate-projects/}{\faExternalLink} & Unsuccessful& Location Available, Exact Site Geometry Unavailable&\\
Arbor Day Foundation \href{https://www.arborday.org/}{\faExternalLink} & Unsuccessful& Location Available, Exact Site Geometry Unavailable&\\
Atlas \href{https://atlas.openforestprotocol.org/}{\faExternalLink} &  Successful& Polygon Geometry Available& API\\
Bôndy \href{https://www.bondy.earth/en/projet}{\faExternalLink} &Unsuccesful&Location mentioned but no geometry available&\\
Climate Action Reserve \href{https://www.climateactionreserve.org/registry/}{\faExternalLink} & Unsuccessful& Could not computationally retrieve data&\\
Climate Impact Partner \href{https://www.climateimpact.com/}{\faExternalLink} & Unsuccessful& Location Available, Exact Site Geometry Unavailable&\\
Climate Partner \href{https://climatepartnerimpact.com/projects/}{\faExternalLink} & Successful& Polygon Geometry Available&\\
Cnaught \href{https://www.cnaught.com/projects}{\faExternalLink} & Unsuccessful& Could not computationally retrieve data&\\
Coeur de Foret \href{https://www.coeurdeforet.com/projets}{\faExternalLink} & Unsuccessful& Location Available, Exact Site Geometry Unavailable&\\
Ecologi \href{https://ecologi.com/projects}{\faExternalLink} & Unsuccessful& Location Available, Exact Site Geometry Unavailable&\\
Ecotree \href{https://ecotree.green/}{\faExternalLink} & Unsuccessful& Location Available, Exact Site Geometry Unavailable&\\
Ecosia \href{https://blog.ecosia.org/tag/where-does-ecosia-plant-trees/}{\faExternalLink} & Unsuccessful& Location Available, Exact Site Geometry Unavailable&\\
Eden People + Planet \href{https://www.edenprojects.org}{\faExternalLink} & Unsuccessful& Location Available, Exact Site Geometry Unavailable&\\
ExplorerLand \href{https://explorer.land/x/projects}{\faExternalLink} &  Successful& Polygon Geometry Available& API\\
Face the Future \href{https://facethefuture.com}{\faExternalLink} & Successful& Polygon Geometry Available& API\\
First Climate \href{https://www.firstclimate.com/klimaschutzprojekte?lang=en}{\faExternalLink} & Unsuccessful& Location Available, Exact Site Geometry Unavailable&\\
Forliance \href{https://forliance.com}{\faExternalLink} & Unsuccessful& Location Available, Exact Site Geometry Unavailable&\\
Forest-trends \href{https://www.forest-trends.org/project-list/#close}{\faExternalLink} & Successful& Point Geometry Available& API\\
Fund Forest (Conservation) \href{https://www.fundforests.org}{\faExternalLink} & Unsuccessful& Location Available, Exact Site Geometry Unavailable&\\
Gold Standard \href{https://registry.goldstandard.org/projects?q=&page=1}{\faExternalLink} & Unsuccessful& Could not computationally retrieve data&\\
Green Climate Fund \href{https://www.greenclimate.fund/projects}{\faExternalLink} & Unsuccessful& Location Available, Exact Site Geometry Unavailable&\\
Greenforestfund \href{https://www.greenforestfund.de/en/sites/}{\faExternalLink} & Successful& Point Geometry Available& Manually Collected\\
GrowMyTree \href{https://growmytree.com/en/pages/transparenz#partnerships}{\faExternalLink} & Unsuccessful& Location Available, Exact Site Geometry Unavailable&\\
Humy \href{https://www.humy.org/nos-projets}{\faExternalLink} & Unsuccessful& Location Available, Exact Site Geometry Unavailable&\\
IDRECCO \href{https://www.reddprojectsdatabase.org/browse-redd-data/}{\faExternalLink} & Unsuccessful& Location Available, Exact Site Geometry Unavailable&\\
Mastreforestation \href{https://www.mastreforest.com/reforestation}{\faExternalLink} & Unsuccessful& Location Available, Exact Site Geometry Unavailable&\\
Mongabay \href{https://reforestation.app/explore?search=\%22\%22&sort=\%22context\%22}{\faExternalLink} & Unsuccessful& Location Available, Exact Site Geometry Unavailable&\\
mossy.earth \href{https://www.mossy.earth/projects}{\faExternalLink} & Unsuccessful& Location Available, Exact Site Geometry Unavailable&\\
One Tree Planted \href{https://onetreeplanted.org/}{\faExternalLink} & Successful& Could not computationally retrieve data&\\
On a mission \href{https://www.onamission.world/projects}{\faExternalLink} & Unsuccessful& Location Available, Exact Site Geometry Unavailable&\\
Plant for Planet \href{https://www.plant-for-the-planet.org/}{\faExternalLink} &  Successful& Polygon Geometry Available& API\\
Primaklima \href{https://www.primaklima.org/was-wir-tun}{\faExternalLink} & Unsuccessful& Location Available, Exact Site Geometry Unavailable&\\
Reforest Action \href{https://www.reforestaction.com/en/projects?category=FOLLOW_UP}{\faExternalLink} & Successful& Point Geometry Available& API\\
Reforestum \href{https://reforestum.com/}{\faExternalLink} &  Successful& Point Geometry Available& API\\
Replant Canada \href{https://www.replant-environmental.ca/nationalparks.html}{\faExternalLink} & Unsuccessful& Could not computationally retrieve data&\\
Restor.eco \href{https://restor.eco/}{\faExternalLink} &  Successful& Polygon Geometry Available& API\\
Terre du Futur \href{https://www.terre-du-futur.fr/projet-de-reboisement-en-france/}{\faExternalLink} & Unsuccessful& Location Available, Exact Site Geometry Unavailable&\\
TIST \href{https://program.tist.org/}{\faExternalLink} & Successful& Polygon Geometry Available&\\
Tree Nation \href{https://tree-nation.com/}{\faExternalLink} &  Successful& Polygon Geometry Available& API\\
Trees for Life \href{https://treesforlife.org.uk/support/plant-a-tree/}{\faExternalLink} & Unsuccessful& Location Available, Exact Site Geometry Unavailable&\\
Trees for the Future \href{https://trees.org/about-us/}{\faExternalLink} & Successful& Point Geometry Available& API\\
Treedom.net \href{https://www.treedom.net/}{\faExternalLink} &  Unsuccessful& Could not computationally retrieve data&\\
United Nations Offset Program \href{https://offset.climateneutralnow.org/reforestation-and-afforestation}{\faExternalLink} & Unsuccessful& Reforestation Data Unavailable&\\
Verra \href{https://registry.verra.org}{\faExternalLink} & Successful& Polygon Geometry Available& Download\\
Verritree \href{https://www.veritree.com/}{\faExternalLink} &  Successful& Point Geometry Available& API\\
Winrock \href{https://acr2.apx.com/myModule/rpt/myrpt.asp?r=111}{\faExternalLink} & Unsuccessful& Could not computationally retrieve data&\\
Zeroco2 \href{https://zeroco2.eco/en/projects/}{\faExternalLink} & Successful& Point Geometry Available& API\\
\hline
\end{tabular}
\caption{List of websites considered}\label{tab:list_websites}

\end{table}

\begin{table}[ht]
\small
\centering
\begin{tabular}{|l|p{7cm}|p{4.5cm}|}
\hline
\textbf{Indicator name} & \textbf{Definition} & \textbf{Data Source} \\
\hline\hline
planting\_date\_reported &
Date when most of the tree planting took place &
Organization metadata \\
\hline
trees\_planted\_reported &
Number of trees planted &
Organization metadata \\
\hline
site\_sqkm &
Size of the site in square kilometres &
Organization metadata / GeoPandas \\
\hline
survival\_rate\_reported &
Share of trees survived (tbd years after planting) &
Organization metadata \\
\hline
species\_planted\_reported &
Tree species planted at each site &
Organization metadata \\
\hline
description\_reported &
Site description provided by project developer &
Organization metadata \\
\hline
project\_id\_reported &
Project ID from project website &
Organization metadata \\
\hline
site\_id\_reported &
Site ID from project website &
Organization metadata \\
\hline
species\_count\_reported &
Number of species reported planted &
Organization metadata \\
\hline
geometry\_reported &
Geo-coordinates provided by project website &
Organization metadata \\
\hline
country &
Country of site location &
Organization metadata / GeoPandas \\
\hline
url &
URL of project organization’s page &
Organization website \\
\hline
host\_name &
Name of organization hosting site info &
Organization website \\
\hline
Nested\_in &
IDs of intersecting polygons &
— \\
\hline
exact\_admin\_area &
Whether polygon aligns with admin area &
GADM \\
\hline
Polygon\_acircle\_oval\_98 &
Whether polygon is (almost) circular/oval &
— \\
\hline
project\_geometries\_invalid &
Whether geometry is valid (polygon/point) &
— \\
\hline
species\_planted\_derived &
Species extracted via LLM from PDFs/descriptions &
Project PDFs / LLM output \\
\hline
planting\_dates\_extracted &
Planting dates extracted via LLM &
Project PDFs / LLM output \\
\hline
\end{tabular}
\caption{Indicators sourced from project websites and derived from PDFs}
\label{tab:metadata_derived}
\end{table}

\begin{table}[ht]
\small
\centering
\begin{tabular}{|l|p{7cm}|p{5cm}|}
\hline
\textbf{Indicator name} & \textbf{Definition} & \textbf{Data Source} \\
\hline\hline
total\_road\_length\_km &
Length of roads within the site (km) &
NASA gROADS (Earth Engine) \\
\hline
built\_area &
Built-up area within the site (sq km) &
GHS BUILT Settlement (Earth Engine) \\
\hline
loss\_pre\_5 &
Average tree loss five years before planting &
GFW Global Forest Change (Earth Engine) \\
\hline
loss\_post\_3 &
Average tree loss three years after planting &
GFW Global Forest Change (Earth Engine) \\
\hline
loss\_post\_5 &
Average tree loss five years after planting &
GFW Global Forest Change (Earth Engine) \\
\hline
cropland\_from\_tree &
Forest-to-cropland conversion area (2000–2020) &
GLAD (Earth Engine) \\
\hline
cropland\_to\_tree &
Cropland-to-forest conversion area (2000–2020) &
GLAD (Earth Engine) \\
\hline
permanent\_water &
Water body area within the site (sq km) &
GLAD (Earth Engine) \\
\hline
short\_vegetation\_after\_tree\_loss &
Re-vegetation after tree loss (sq km, 2000–2020) &
GLAD (Earth Engine) \\
\hline
top\_three\_ndvi\_months &
Top three months with highest NDVI in 2023 &
Sentinel-2 (Earth Engine) \\
\hline
tree\_cover\_area\_2020 &
Tree cover area in 2020 (sq km) &
GLAD (Earth Engine) \\
\hline
tree\_cover\_area\_2015 &
Tree cover area in 2015 (sq km) &
GLAD (Earth Engine) \\
\hline
tree\_cover\_area\_2010 &
Tree cover area in 2010 (sq km) &
GLAD (Earth Engine) \\
\hline
tree\_cover\_area\_2005 &
Tree cover area in 2005 (sq km) &
GLAD (Earth Engine) \\
\hline
tree\_cover\_area\_2000 &
Tree cover area in 2000 (sq km) &
GLAD (Earth Engine) \\
\hline 
average\_precipitation &
Average monthly rainfall at centroid (planting, 1, 2, 5 yr) &
WorldClim \cite{harris2020version} \\
\hline
tmax\_and\_tmin &
Max/min temperature at planting, 1, 2, 5 yr &
WorldClim \cite{harris2020version} \\
\hline
Change\_1year &
Tree-cover change (NDVI + Shadow Index) 1 yr post-planting &
COPERNICUS/S2\_HARMONIZED \\
\hline
Change\_2years &
Tree-cover change (NDVI + Shadow Index) 2 yr post-planting &
COPERNICUS/S2\_HARMONIZED \\
\hline
Change\_5years &
Tree-cover change (NDVI + Shadow Index) 5 yr post-planting &
COPERNICUS/S2\_HARMONIZED \\
\hline
mean\_elevation &
Mean elevation of site (m) &
NASA SRTM DEM (30 m) \\
\hline
mean\_slope &
Mean slope of site &
NASA SRTM DEM (30 m) \\
\hline
\end{tabular}
\caption{Indicators from secondary spatial and environmental datasets}
\label{tab:secondary_data}

\end{table}

\end{document}